\documentclass[onecolumn,10pt]{article}
\usepackage[top=1.25in, bottom=1.25in, left=1.25in, right=1.25in]{geometry}
\usepackage{amsmath,amsfonts,amscd,amssymb}
\usepackage{graphicx}
\usepackage{algorithm, algorithmic}

\usepackage{overpic}
\usepackage{cancel}
\usepackage{rotating}
\usepackage{url}
\usepackage{caption}
\usepackage{color}
\usepackage{rotating}
\usepackage{multirow}
\usepackage{wrapfig}
\usepackage{mathtools}
\usepackage{subeqnarray}
\usepackage{setspace}
\usepackage{hyperref}
\usepackage{palatino} 
\usepackage[bottom,flushmargin,hang,multiple]{footmisc}

\newcommand\blfootnote[1]{%
  \begingroup
  \renewcommand\thefootnote{}\footnote{#1}%
  \addtocounter{footnote}{-1}%
  \endgroup
}

\renewenvironment{abstract}
 {\par\noindent\rule{\linewidth}{.25pt}\par\noindent\textbf{\abstractname}\par\noindent \ignorespaces}
 {\par\noindent\medskip\rule{\linewidth}{.25pt}}
 
\DeclareMathOperator{\diag}{diag}

\newcommand{\bx}{\mathbf{x}}
\newcommand{\by}{\mathbf{y}}
\newcommand{\be}{\mathbf{e}}

\newcommand{\bc}{\mathbf{c}}

\newcommand{\bC}{\mathbf{C}}

\newcommand{\bX}{\mathbf{X}}
\newcommand{\bI}{\mathbf{I}}

\newcommand{\bbR}{\mathbb{R}}

\newcommand{\cN}{\mathcal{N}}
\newcommand{\cD}{\mathcal{D}}

\newcommand{\brho}{\boldsymbol{\rho}}
\newcommand{\bmu}{\boldsymbol{\mu}}

\newcommand{\bxi}{\boldsymbol{\xi}}
\newcommand{\bnu}{\boldsymbol{\nu}}
\newcommand{\bgamma}{\boldsymbol{\gamma}}
\newcommand{\bGamma}{\boldsymbol{\Gamma}}
\newcommand{\bSigma}{\boldsymbol{\Sigma}}
\newcommand{\bTheta}{\boldsymbol{\Theta}}
\newcommand{\obTheta}{\overline{\boldsymbol{\Theta}}}
\newcommand{\btheta}{\boldsymbol{\theta}}

\usepackage[]{lineno}

\usepackage{titlesec}
\titleformat*{\section}{\bfseries}
\titleformat*{\subsection}{\itshape}
\titleformat*{\subsubsection}{\itshape}
\titleformat*{\paragraph}{\large\bfseries}
\titleformat*{\subparagraph}{\large\bfseries}

\title{\Large{\vspace{-.65in}{Sparse Methods for Automatic Relevance Determination}}\vspace{-.25in}}
\author{\normalsize{Samuel H. Rudy$^{1*}$, Themistoklis P. Sapsis$^1$} \\ 
\footnotesize{$^1$Department of Mechanical Engineering, Massachusetts Institute of Technology, Cambridge, MA 02139} \vspace{-1 mm}
}
\date{}
\begin{document}
\maketitle

\blfootnote{$^*$ Corresponding author (shrudy@mit.edu).\\ \noindent \textbf{Python code:}  \href{https://github.com/snagcliffs/SparseARD}{https://github.com/snagcliffs/SparseARD}}
\vspace{-.25in}
\begin{abstract}
This work considers methods for imposing sparsity in Bayesian regression with applications in nonlinear system identification.  
We first review automatic relevance determination (ARD) and analytically demonstrate the need to additional regularization or thresholding to achieve sparse models.  
We then discuss two classes of methods, regularization based and thresholding based, which build on ARD to learn parsimonious solutions to linear problems.  
In the case of orthogonal covariates, we analytically demonstrate favorable performance with regards to learning a small set of active terms in a linear system with a sparse solution.
Several example problems are presented to compare the set of proposed methods in terms of advantages and limitations to ARD in bases with hundreds of elements.
The aim of this paper is to analyze and understand the assumptions that lead to several algorithms and to provide theoretical and empirical results so that the reader may gain insight and make more informed choices regarding sparse Bayesian regression. \\

\vspace{-0.05in}
\vspace{-0.05in}
\end{abstract}



\section{Introduction}\label{sec:intro}

%
In many modeling and engineering problems it is critical to build statistical models from data which include estimates of the model uncertainty.  
This is often achieved through non-parametric Bayesian regression in the form of Gaussian processes and similar methods \cite{rasmussen2003gaussian}.
While these methods offer tremendous flexibility and have seen success in a wide variety of applications they have two significant shortcomings; they are not interpretable and they often fail in high dimensional settings. 
The simplest case of parametric Bayesian regression is Bayesian ridge regression, where one learns a distribution for model parameters by assuming identical independently distributed (iid) Gaussian priors on model weights.
However, Bayesian ridge requires the researcher to provide a single length scale for the prior that stays fixed across dimensions.  It is therefore not invariant to changes in units.
Furthermore Bayesian ridge regression does not yield sparse models.  
In a high dimensional setting this may hinder the interpretability of the learned model.
Automatic Relevance Determination (ARD) \cite{neal2012bayesian, tipping2001sparse, wipf2008new} addresses both of these problems.  ARD learns length scales associated with each free variable in a regression problem.
In the context of linear regression, ARD is often referred to as Sparse Bayesian Learning (SBL) \cite{tipping2003fast,wipf2008new,wu2012dual} due to its tendency to learn sparse solutions to linear problems. 
ARD has been applied to problems in 
compressed sensing \cite{babacan2009bayesian},
sparse regression \cite{wipf2004basis, zhang2018robust, zhang2019robust, yuan2019machine, fuentes2019efficient},
matrix factorization, \cite{tan2012automatic},
classification of gene expression data \cite{li2002bayesian},
earthquake detection \cite{oh2008bayesian},
Bayesian neural networks \cite{neal2012bayesian}, as well as other fields.

More recently, some works have used ARD for interpretable nonlinear system identification. 
In this setting a linear regression problem is formulated to learn the equations of motion for a dynamical system from a large collection of candidate functions, called a library.
Traditionally, frequentist methods have been applied to select a small set of active terms from the candidate functions.
These include symbolic regression \cite{Bongard2007pnas, Schmidt2009science, quade2016prediction}, sequential thresholding \cite{Brunton2016,rudy2019data,schaeffer2017learning}, information theoretic methods \cite{almomani2020entropic, kim2017causation}, relaxation methods, \cite{tibshirani1996regression, zheng2018unified}, and constrained sparse optimization \cite{tran2017exact}.
Bayesian methods for nonlinear system identification \cite{yang2020bayesian, niven2020bayesian} including ARD have been applied to the nonlinear system identification problem for improved robustness in the case of low data \cite{yuan2019machine} and for uncertainty quantification \cite{zhang2018robust, zhang2019robust, fuentes2019efficient}.
The critical challenge for any library method for nonlinear system identification is learning the correct set of active terms.
%

%

Motivated by problems such as nonlinear system identification, where accuracy in determining the sparsity pattern on a predictor is paramount, we focus on the ability of ARD to accurately learn a small subset of active terms in a linear system.
This is in contrast to convergence in the sense of any norm.
Indeed, it was previously shown \cite{yuan2019machine} that ARD converges to the true predictor as noise present in the training data shrinks to zero.
However, we show analytically that ARD fails to obtain the true sparsity pattern in the case of an orthonormal design matrix, leading to extraneous terms for arbitrarily small magnitudes of noise.
This result motivates further considerations for imposing sparsity on the learned model.

This paper explores several intuitive methods for imposing sparsity in the ARD framework.
We discuss the assumptions that lead to each technique, any approximations we use to make them tractable, and in some cases provide theoretical results regarding their accuracy with respect to selection of active terms.
We stress that while sparse regression is a mature field with many approaches designed to approximate the $\ell^0$-penalized least squares problem \cite{tibshirani1996regression,zhang2019convergence, almomani2020entropic, zhang2009greedy, blumensath2009iterative}, most of these techniques do not consider uncertainty.  
We therefore only compare results of the proposed techniques to ARD.

The paper is organized as follows.  In section \ref{sec:setup} we provide a brief discussion on the automatic relevance determination method for Bayesian linear regression. Section \ref{sec:regularization_methods} introduces two regularization-based methods for imposing sparsity of learned predictors from the ARD algorithm.  Section \ref{sec:thresholding_methods} introduces various thresholding-based approaches.  In each case we provide analytical results for the expected false positive and negative rates with respect to coefficients being set to zero.  Section \ref{sec:comparison} includes a more detailed comparison between methods.  Section \ref{sec:results} includes results of each of the proposed methods applied to a variety of problems including a sparse linear system, function fitting, and nonlinear system identification.  Discussion and comments towards future work are included in Section \ref{sec:discussion}.

\section{Setup}\label{sec:setup}

We start with the likelihood model,
\begin{equation}
    \begin{aligned}
        y &= \btheta(\bx)\bxi + \nu \\
        \nu &\sim \cN(0, \sigma^2),
    \end{aligned}
    \label{eq:sparse_reg}
\end{equation}
where $\btheta: \bbR^n \to \bbR^d$ forms a nonlinear basis, $y$ is scalar, $\bx\in \bbR^{n}$, $\bxi \in \bbR^d$ and $\nu$ is normally distributed error with variance $\sigma^2$.  We assume a prior distribution on weights $\xi$ with variance given by hyper-parameter $\bgamma$.
\begin{equation}
    \xi_i \sim \cN (0, \gamma_i)
    \label{eq:xi_prior}
\end{equation}
Automatic relevance determination seeks to learn the value of parameter $\bgamma$ that maximizes evidence.  This approach is known as evidence maximization, empirical Bayes, or type-II maximum likelihood \cite{bishop2006pattern, murphy_ml}.  Given a dataset $\cD = \{(\bx_i ,y_i)\}_{i=1}^m$ we marginalize over $\bxi$ to obtain the posterior likelihood of $\bgamma$.  This gives,
\begin{equation}
\begin{aligned}
    p(\cD | \bgamma) &= \int p(\cD | \bxi) p(\bxi ; \bgamma) \, d\bxi \\
    &\propto |\bSigma_y|^{-\frac{1}{2}} exp\left( -\frac{1}{2} \by^T \bSigma_y^{-1} \by \right),
    \end{aligned}
    \label{eq:gamma_posterior}
\end{equation}
\noindent where $\bSigma_y = \sigma^2 \bI_m + \btheta(\bX)\bGamma\btheta(\bX)^T$, $\bGamma = diag(\bgamma)$, $\by$ is a column vector of all observed outputs and $\btheta(\bX)\in \bbR^{d \times m}$ is a matrix whose rows are the nonlinear features of each observed $\bx$. We estimate $\bgamma$ by maximizing Eq. \eqref{eq:gamma_posterior} and the subsequent distribution for $\bxi$, given by $\bxi \sim \mathcal{N}(\bmu_\xi, \bSigma_\xi)$.  Letting $\bTheta = \btheta(\bX)$ this is,
\begin{equation}
    \begin{aligned}
        \bmu_{\xi } &=  \sigma^{-2} \bSigma_\xi \bTheta^T \by \\
        \bSigma_{\xi } &= \left(\sigma^{-2} \bTheta^T \bTheta + \bGamma^{-1} \right)^{-1}
    \end{aligned}
    \label{eq:xi_posterior}
\end{equation}
In practice $\bgamma$ is found by minimizing the negative log of Eq. \eqref{eq:gamma_posterior} given by,
\begin{equation}
    L(\bgamma) = -\log p(\cD ; \bgamma) \propto log |\bSigma_y| + \by^T \bSigma_y^{-1} \by .
    \label{eq:neg_log_likelihood}
\end{equation}
Following \cite{yuan2019machine} (see Appendix A) the second term in \eqref{eq:neg_log_likelihood} is equivalent to,
\begin{equation}
\by^T \bSigma_y^{-1} \by = \underset{\bxi}{min} \frac{1}{\sigma^2} \|\by - \bTheta \bxi \|_2^2 + \bxi^T\bGamma^{-1}\bxi , \label{eq:convex_loss_equiv}
\end{equation}
which gives the following representation of the loss function \eqref{eq:neg_log_likelihood},
\begin{equation}
L(\bgamma) = \underset{\bxi}{min} \, \left(log |\bSigma_y| + \frac{1}{\sigma^2} \|\by - \btheta \bxi \|_2^2 + \bxi^T\bGamma^{-1}\bxi  \right) .
    \label{eq:aux_neg_log_likelihood}
\end{equation}
To minimize \eqref{eq:aux_neg_log_likelihood} we solve a sequence of $\ell^1$-penalized least squares problems developed in \cite{wipf2008new}.  This is shown in Alg.~\ref{alg:ARD}.

\begin{algorithm}[H]
\begin{algorithmic}[1]
\STATE \vspace{1 mm} Initialize $\bgamma$
\STATE \vspace{1 mm} while not converged:
\STATE \vspace{1 mm} \hspace{5 mm}$\bc^{(k+1)} = \nabla_{\bgamma} \left( \log \left| \bSigma_y^{(k)} \right| \right) = diag\left(  \bTheta^T{\bSigma_y^{(k)}}^{-1} \bTheta \right)  $\\ 
\hspace{5 mm} where $\bSigma_y^{(k)} = \sigma^2 \bI + \bTheta \bGamma^{(k)} \bTheta $
\STATE \vspace{1 mm} \hspace{5 mm}  $\bxi^{(k+1)} =  \underset{\bxi}{arg min} \left\{ \|\by - \bTheta \bxi \|^2 + \sum_i \eta_i^{(k+1)}| \xi_i | \right\}$ \\ 
\hspace{5 mm} where $\eta_i^{(k+1)} = 2 \sigma^2 \sqrt{c_i^{(k+1)}}$
\STATE \vspace{1 mm} \hspace{5 mm} $\gamma^{(k+1)}_i = {c_i^{(k+1)}}^{-1/2} |\xi_i^{(k+1)}| $ \\
\STATE \vspace{1 mm} \hspace{5 mm} Optional: relearn $\sigma^2$
\STATE \vspace{1 mm}return $\bgamma^{(k+1)}$
\end{algorithmic}
\caption{ARD($\bTheta , \by, \sigma^2$)}
\label{alg:ARD}
\end{algorithm}

Some works have used Gamma distribution priors on scale parameters $\bgamma$ and precision $\sigma^{-2}$ \cite{tipping2001sparse}.  This leads to a problem that is solved using coordinate descent of a slightly altered loss function from that shown in Eq.~\eqref{eq:aux_neg_log_likelihood}.  More recent works \cite{wipf2008new,yuan2019machine,zhang2018robust} have not used this formulation, so much of the following work does not use hierarchical priors.  We note however, that the case of a Gamma distribution prior on $\bgamma$ with shape parameter $k=1$ is in fact a Laplace prior.  This case has been studied as a Bayesian compressed sensing method \cite{babacan2009bayesian} and is a special case of the formulation considered in Sec. \ref{sec:regularization_methods}.

The minimization problem in step 4 of Alg.~\ref{alg:ARD} may be re-written, after rescaling $\bTheta$ and $\bxi$, to obtain the commonly used Lagrangian form of the least absolute shrinkage and selection operator (Lasso) \cite{tibshirani1996regression}.  Letting,
\begin{equation}
\boldsymbol{\zeta}^{(k+1)} = \underset{\boldsymbol{\zeta}}{arg min} \left\| \by - \bTheta \, diag\left( \boldsymbol{\eta}^{(k+1)} \right)^{-1} \boldsymbol{\zeta} \right\|_2^2 + \|\boldsymbol{\zeta}\|_1
\label{eq:lasso_normal}
\end{equation}
we get,
\begin{equation}
\bxi^{(k+1)} = diag\left( \boldsymbol{\eta}^{(k+1)} \right)^{-1} \boldsymbol{\zeta}.
\end{equation}
Typical solvers for Eq. \eqref{eq:lasso_normal} include coordinate descent \cite{wu2008coordinate}, proximal gradient methods \cite{parikh2014proximal}, alternating direction method of multipliers \cite{boyd2011distributed}, and least angle regression (LARS) \cite{efron2004least}.  Several example datasets considered in this manuscript resulted in ill-conditioned $\bTheta$ and therefore slow convergence of algorithms for solving the Lasso subroutine.  We found empirically that all methods performed equally well on orthogonal $\bTheta$ but for ill-conditioned cases LARS far outperformed other optimization routines.

As we have noted, it is often the case that solutions to Eq. \eqref{eq:neg_log_likelihood} exhibit some degree of sparsity.  
However, such solutions are only sparse in comparison to those derived by methods such as Bayesian ridge, where all coefficients are nonzero.  
For problems where we seek to find a very small set of nonzero terms, Alg. \ref{alg:ARD} must be adjusted to push extraneous terms to zero.  
In the following two sections we will discuss five methods for doing so.

\section{Regularization Based Methods}\label{sec:regularization_methods}

We begin with a discussion of two methods for regularizing ARD to obtain more sparse predictors: inflating the variance passed into Alg.~\ref{alg:ARD} and including a prior for the distribution of $\bgamma$.  In each case the sparse predictor is found as the fixed point of an iterative algorithm.  In subsequent sections we will discuss thresholding based methods that alternate between iterative optimization and thresholding operations.  In certain cases we refer to the set valued subgradient of a continuous piecewise differentiable function.  In cases where the subgradient is a singleton we treat it as a real number.

\subsection{Variance Inflation}\label{subsec:ARDvi}

The error variance $\sigma^2$ of likelihood model \eqref{eq:sparse_reg} may be intuitively thought of as a level of mistrust for the data $\cD$.  Extremely large values of $\sigma^2$ will push estimates of $\bxi$ to be dominated by priors.  It is shown in \cite{wipf2008new} that the ARD prior given by \eqref{eq:xi_prior} is equivalent to a concave regularization term.  We therefore expect large $\sigma^2$ to encourage more sparse models.  The regularization may be strengthened by passing in an artificially large value of $\sigma^2$ to the iterative algorithm for solving Eq.~\eqref{eq:aux_neg_log_likelihood} or, if also learning $\sigma^2$, by applying an inflated value at each step in the algorithm.  We will call this process ARD with variance inflation (ARDvi), shown in Algorithm \ref{alg:ARDvi}.  Note that this differs from Alg. \ref{alg:ARD} only slightly by treating the variance used in the standard ARD algorithm as a tuning parameter, with a higher variance indicating less trust in the data and a greater regularization.

\begin{algorithm}[]
\begin{algorithmic}[1]
\STATE \vspace{1 mm} Initialize $\bgamma$
\STATE \vspace{1 mm} while not converged:
\STATE \vspace{1 mm} \hspace{5 mm}$\bc^{(k+1)} = \nabla_{\bgamma} \left( \log \left| \bSigma_y^{(k)} \right| \right) = diag\left(  \bTheta^T{\bSigma_y^{(k)}}^{-1} \bTheta \right)  $\\ 
\hspace{5 mm} where $\bSigma_y^{(k)} = \alpha\sigma^2 \bI + \bTheta \bGamma^{(k)} \bTheta$
\STATE \vspace{1 mm} \hspace{5 mm}  $\bxi^{(k+1)} =  \underset{\bxi}{arg min} \left\{ \|\by - \bTheta \bxi \|^2 + \sum_i \eta_i^{(k+1)}| \xi_i | \right\}$ \\ 
\hspace{5 mm} where $\eta_i^{(k+1)} = 2 \alpha\sigma^2 \sqrt{c_i^{(k+1)}}$
\STATE \vspace{1 mm} \hspace{5 mm} $\gamma^{(k+1)}_i = {c_i^{(k+1)}}^{-1/2} |\xi_i^{(k+1)}| $ \\
\STATE \vspace{1 mm} \hspace{5 mm} Optional: relearn $\sigma^2$
\STATE \vspace{1 mm}return $\bgamma^{(k+1)}$
\end{algorithmic}
\caption{ARDvi($\bTheta , \by, \sigma^2, \alpha=$ inflation factor)}
\label{alg:ARDvi}
\end{algorithm}

\subsubsection{Sparsity properties of ARDvi for orthogonal features}

To better understand the effect of variance inflation we consider Alg.~\ref{alg:ARDvi} in the case where columns of $\bTheta$ are othogonal.  Note that this implies $m \geq n$.  Let $\sqrt{\brho}$ be the vector of column norms of $\bTheta$ so that $\bTheta^T\bTheta = diag(\brho)$.  Define $\overline{\bTheta}$ to be the extension of $\bTheta$ to an orthogonal basis of $\bbR^m$ that $\overline{\bTheta}^T \overline{\bTheta} = \textbf{R} = diag(\overline{\brho})$ with the first $n$ entries of $\overline{\brho}$ given by $\brho$.  Now let $\bgamma^*$ be a fixed point of algorithm \ref{alg:ARDvi}, $\overline{\bGamma}^* = diag(\bgamma^*, \textbf{0}_{m-n}) \in \bbR^{m \times m}$, and $\bc^*$, $\bxi^*$ be defined by steps 3 and 4. The expression in step 3 is given by,
\begin{equation}
\begin{aligned}
c_i^* &= \bTheta_i^T \left( \alpha\sigma^2\bI + \bTheta \bGamma^*\bTheta^T \right)^{-1}\bTheta_i \\
&= \bTheta_i^T \left( \alpha\sigma^2 \obTheta \textbf{R}^{-1} \obTheta^T  + \obTheta \overline{\bGamma}^*\obTheta^T \right)^{-1}\bTheta_i \\
&= \bTheta_i^T \left(\obTheta \left(\alpha\sigma^2 \textbf{R}^{-1} + \overline{\bGamma}^* \right)\obTheta^T \right)^{-1}\bTheta_i \\
&= \bTheta_i^T \obTheta \textbf{R}^{-1} \left(\alpha\sigma^2 \textbf{R}^{-1} + \overline{\bGamma}^* \right)^{-1} \textbf{R}^{-1} \obTheta^T \bTheta_i \\
&= \be_i^T \left(\alpha\sigma^2 \textbf{R}^{-1} + \overline{\bGamma}^* \right)^{-1} \be_i \\
&= \dfrac{1}{\alpha\sigma^2\rho_i^{-1} + \bgamma_i^* }  \\
&= \dfrac{1}{\alpha\sigma^2\rho_i^{-1} + \frac{|\xi_i^*|}{\sqrt{c_i^*}}}  \\
 \sqrt{c_i^*} &= \dfrac{-|\xi_i^*| + \sqrt{{\xi_i^*}^2 + 4 \alpha\sigma^2\rho_i^{-1}}}{2\alpha\sigma^2\rho_i^{-1}}
\end{aligned} \label{eq:root_c*}
\end{equation}
The Karush-Kugn-Tucker (KKT) stationarity condition for the $\xi$ update in step 4 gives,
\begin{equation}
\begin{aligned}
0 &\in \bTheta_i^T\left(  \bTheta\bxi^* - \by  \right) + \alpha\sigma^2 \sqrt{c_i^*} \partial |\xi_i^*| \\
&\in \bTheta_i^T\left(  \bTheta\xi^* - (\bTheta \bxi + \bnu) \right) + \alpha\sigma^2 \sqrt{c_i^*} \partial |\xi_i^*| \\
&\in \rho_i\xi^*_i - \rho_i\xi_i + \bTheta_i^T\bnu+ \alpha\sigma^2 \sqrt{c_i^*} \partial |\xi_i^*|,
\end{aligned}
\end{equation}
where $\bxi$ denotes the true value from Eq. \eqref{eq:sparse_reg}.  We can find the false positive probability for a term being included in the model by setting $\xi_i = 0$ and finding conditions under which $\xi^*_i \neq 0$.  Subbing in the value for $\sqrt{c_i^*}$ from Eq. \eqref{eq:root_c*} and dividing by $\rho_i$ gives,
\begin{equation}
\begin{aligned}
\xi_i = 0 \Rightarrow  -\rho_i^{-1}\bTheta_i^T\bnu \in \xi_i^* + \dfrac{1}{2} \left( \sqrt{{\xi_i^*}^2 + 4 \alpha\sigma^2\rho_i^{-1}}-|\xi_i^*|\right) \partial |\xi_i^*|,
\end{aligned}
\end{equation}
where $\partial |\xi_i^*|$ is a set valued function taking value $[-1,1]$ if $\xi_i^*=0$ or $\{ \text{sgn}(\xi_i^*) \}$ otherwise.  If $\xi_i^* = 0$ then,
\begin{equation}
|\rho_i^{-1}\bTheta_i^T\bnu| \leq \underset{z \in \partial |\xi_i^*|}{\text{sup}} \left| z \sqrt{\alpha \sigma^2\rho_i^{-1}}\right| = \sqrt{\alpha \sigma^2\rho_i^{-1}},
\end{equation}
while for $\xi^* \neq 0$,
\begin{equation}
\begin{aligned}
|\rho_i^{-1}\bTheta_i^T\bnu| &= \left| \xi_i^* + \dfrac{1}{2} \left( \sqrt{{\xi_i^*}^2 + 4 \alpha\sigma^2\rho_i^{-1}}-|\xi_i^*|\right) sgn(\xi_i^*) \right| \\
&=  \dfrac{1}{2} \left( \left| \xi_i^* \right| + \sqrt{{\xi_i^*}^2 + 4 \alpha\sigma^2\rho_i^{-1}}  \right) \\
&> \sqrt{\alpha \sigma^2\rho_i^{-1}}.
\end{aligned}
\end{equation}
It follows that $p(\xi_i^* \neq 0 | \xi_i = 0) = p(\rho_i^{-1}|\bTheta_i^T\bnu| > \sqrt{\alpha \sigma^2\rho_i^{-1}})$.  Since $\rho_i^{-1} \bTheta_i^T\bnu \sim \mathcal{N}(0, \rho_i^{-1} \sigma^2)$ we find that the false positive rate is,
\begin{equation}
FP_{VI}(\alpha) = p(\xi_i^* \neq 0 | \xi_i = 0) = 1 - \text{erf} \left( \sqrt{ \frac{\alpha}{2}  } \right),
\label{eq:VI_FP_rate}
\end{equation}
where $erf$ is the Gauss error function.  Of particular note is that the number of false positives is independent from the variance of the linear model's error term, $\sigma^2$.  While the mean predictor learned from ARD does converge in any norm to the true solution for $\sigma^2 \to 0$, the expected number of nonzero terms in the learned predictor stays constant.  If one desires a sparse predictor, this motivates including a small threshold parameter below which coefficients are ignored, which we will discuss in a subsequent section.  

We define a false negative by Algorithm \ref{alg:ARDvi} inding some $\gamma^*_i = 0$ (and respectively $\xi_i^*$) when the true solution $\xi_i \neq 0$ and find the likelihood of such a case in a similar manner. Applying Eq. \eqref{eq:root_c*} and the KKT conditions we find,
\begin{equation}
\begin{aligned}
0 &\in -\xi_i + \rho_i^{-1}\bTheta_i^T\bnu + \sqrt{\alpha\sigma^2\rho_i^{-1}} sgn(\xi_i^*) \\
\rho_i^{-1}\bTheta_i^T\bnu &\in \xi_i - \sqrt{\alpha\sigma^2\rho_i^{-1}} sgn(\xi_i^*) \\
& \in \left[ \xi_i - \sqrt{\alpha \sigma^2\rho_i^{-1}} ,  \xi_i + \sqrt{\alpha \sigma^2\rho_i^{-1}} \right]
\end{aligned}
\end{equation}
The false negative likelihood is therefore,
\begin{equation}
FN_{VI}(\alpha) = p(\xi_i^* = 0 | \xi_i \neq 0) = \frac{1}{2} \left( \text{erf}\left( \frac{\xi_i + \sqrt{\alpha \sigma^2\rho_i^{-1}}}{\sigma \sqrt{2 \rho_i^{-1}}} \right) - \text{erf}\left( \frac{\xi_i - \sqrt{\alpha \sigma^2\rho_i^{-1}}}{\sigma \sqrt{2 \rho_i^{-1}}} \right)  \right)
\label{eq:VI_FN_rate}
\end{equation}
Note that this function vanishes for large $|\xi_i|$, indicating that important terms, as measured by $|\xi_i|$ are far less likely to be missed.

\begin{figure}
\centering
\includegraphics[width=\textwidth]{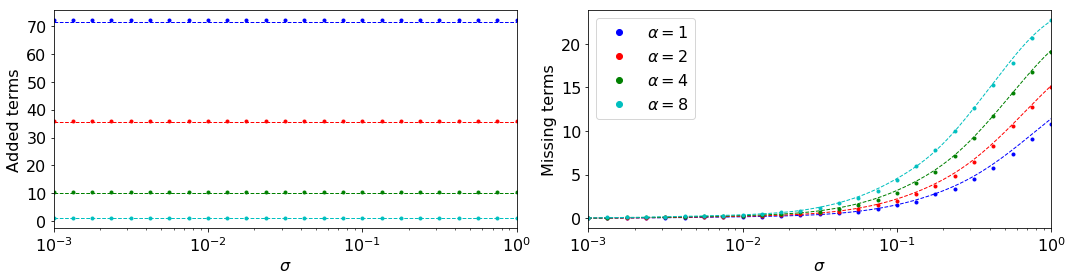}
\caption{Number of missing and added terms using ARD with variance inflation.  Dots indicate empirical average over 50 trials and dashed lined indicate values predicted by equations \eqref{eq:VI_FP_rate} and \eqref{eq:VI_FN_rate}.  Since theoretical number of missing terms is dependent on $\xi$, it was held fixed across all trials and only noise $\nu$ was sampled. In each case $\bTheta$ is a 250x250 matrix with orthogonal columns.}
\label{fig:orthonormal_variance_inflation}
\end{figure}

Figure \ref{fig:orthonormal_variance_inflation} demonstrates the validity of equations \eqref{eq:VI_FP_rate} and \eqref{eq:VI_FN_rate} on a simple test problem.  We construct a matrix $\bTheta \in \bbR^{250 \times 250}$ with orthogonal columns having random magnitude such that $\rho_i \sim \mathcal{U}([1,3])$ and random $\bxi$ with $\|\bxi\|_0 = 25$ having non-zero terms distributed according to $\cN(0,1)$.  The mean number of added and missed nonzero terms across 50 trials are shown and agree very well with the predicted values.  As anticipated, the number of missing terms decays to zero as $\sigma \to 0$, but the same is not true for the number of added terms, which only decays as the inflation parameter $\alpha$ is increased.  The failure of ARDvi to converge as $\sigma \to 0$ to the true sparsity pattern for fixed $\alpha$ is certainly troubling, but for sufficiently large $\alpha$ only an arbitrarily small number of terms will be added.

\subsection{Regularization via Sparsity Promoting Hierarchical Priors}\label{subsec:ARDr}

Since the sparsity of $\bxi$ is controlled by that of $\bgamma$ we can also attempt to regularize $\bgamma$ through the use of a hierarchical prior.
Previous approaches to ARD have suggested hierarchical priors on $\bgamma$ in the form of Gamma distributions \cite{tipping2001sparse}.  However, except for certain cases, the general class of Gamma distributions does not impose sparsity.  Instead, we consider the use of a sparsity promoting hierarchical prior on the scale parameters $\bgamma$.  We consider distributions of the form,
\begin{equation}
    p(\gamma_i) \propto exp \left( \frac{-g(\gamma_i) - f(\gamma_i)}{2} \right),
    \label{eq:gamma_prior}
\end{equation}
where $f$, $g$ are each convex and concave functions in $\gamma_i$, respectively.  Given data $\mathcal{D}$ we can follow a procedure similar to the one used in Sec. \ref{sec:setup} and find,
\begin{equation}
\begin{aligned}
    p(\bgamma | \cD) &\propto p(\cD | \bgamma) p(\bgamma) = \int p(\cD | \bxi) p(\bxi | \bgamma) \, d\bxi \, p(\bgamma) \\
    &= (2\pi)^{-m/2} |\bSigma_y|^{-\frac{1}{2}} exp\left( -\frac{1}{2} \by^T \bSigma_y^{-1} \by \right) \prod_{i=1}^d e^{\left( \frac{-g(\gamma_i) - f(\gamma_i)}{2} \right)}.
    \end{aligned}
    \label{eq:ARDr_gamma_posterior}
\end{equation}
A fully Bayesian approach would estimate $\btheta$ through the joint posterior likelihood of pairs $\btheta, \bgamma$, but this would be computationally expensive.  Instead, we approximate $\bgamma$ by its maximum-a-posteriori estimate $\bgamma_{MAP} = argmax \, p(\bgamma | \cD)$, a process sometimes labelled type-II MAP \cite{murphy_ml}.
The MAP estimate of $\bgamma$ is found by minimizing the negative log of the posterior distribution,
\begin{equation}
\begin{aligned}
    L_{ARDr}(\bgamma) &= -\log p(\cD ; \bgamma) \propto log |\bSigma_y| + \by^T \bSigma_y^{-1} \by + \sum_{i=1}^d \left( f(\gamma_i) + g(\gamma_i) \right) \\
    &= \underset{\bxi}{min} \, \left(log |\bSigma_y| + \frac{1}{\sigma^2} \|\by - \bTheta \bxi \|_2^2 + \bxi^T\bGamma^{-1}\bxi + \sum_{i=1}^d \left( f(\gamma_i) + g(\gamma_i) \right) \right) .
    \end{aligned}
    \label{eq:ARDr_neg_log_likelihood}
\end{equation}
As expected, Eq. \eqref{eq:ARDr_neg_log_likelihood} closely resembles Eq. \eqref{eq:aux_neg_log_likelihood} and may be solved with a similar method.  Alg.~\ref{alg:ARDr} constructs a sequence $\gamma^{(i)}$ which monotonically increases the likelihood given by Eq. \eqref{eq:ARDr_neg_log_likelihood}.  Since $\mathcal{L}_{ARDr}$ is nonconvex, we can only guarantee convergence to a local minimum.  We initialize Alg.~\ref{alg:ARDr} using the unregularized ARD value of $\gamma$.

\begin{algorithm}[H]
\begin{algorithmic}[1]
\STATE \vspace{1 mm} Initialize $\bgamma$ using Algorithm \ref{alg:ARD}
\STATE \vspace{1 mm} while not converged:
\STATE \vspace{1 mm} \hspace{5 mm}$\bc^{(k+1)} = \nabla_{\bgamma} \left( \log \left| \bSigma_y^{(k)} \right| + \sum g\left(\gamma_i^{(k)} \right)  \right)$\\ 
\STATE \vspace{1 mm} \hspace{5 mm}$\bc^{(k+1)} = diag\left(  \bTheta^T{\bSigma_y^{(k)}}^{-1} \bTheta \right) - \nabla_{\bgamma} \sum g\left(\gamma_i^{(k)} \right) $\\ 
\hspace{5 mm} where $\bSigma_y^{(k)} = \sigma^2 \bI + \bTheta \bGamma^{(k)} \bTheta $
\STATE \vspace{1 mm} \hspace{5 mm} $\bgamma^{(k+1)} = \underset{\bgamma}{\text{arg min}} \left\{  \underset{\bxi}{\min} \left\{ \frac{1}{\sigma^2} \|\by - \bTheta \bxi \|^2 + \sum_i \left( \frac{\xi_i^2}{\gamma_i} + c_i^{(k+1)} \gamma_i + f(\gamma_i) \right) \right\} \right\}$  \\
\STATE \vspace{1 mm} \hspace{5 mm} Optional: relearn $\sigma^2$
\STATE \vspace{1 mm}return $\bgamma^{(k+1)}$
\end{algorithmic}
\caption{ARDr($\bTheta, \by, \sigma^2, f, g$)}
\label{alg:ARDr}
\end{algorithm}

Algorithm \ref{alg:ARDr} allows for significant freedom in choosing $f$ and $g$.  The concave component of the prior, $g$, acts as a sparsity encouraging regularizer on $\bgamma$, as is common for concave priors \cite{fan2001variable}.  Examples of concave $g$ include the identity, $tanh$, and approximations of the $\ell^0$-norm.  We consider functions of the following form;
\begin{equation}
\begin{aligned}
g_{\lambda, \eta}(\gamma_i) &= \text{min}\{\lambda \gamma_i, \eta \} \\
\end{aligned}
\label{eq:g_examples}
\end{equation}
where $\lambda$ is a parameter controlling the strength of the regularizer and $\eta$ is a width parameter. The convex prior $f$ may an indicator function restricting $\gamma$ to a specific domain or left as a constant.  In either case implementing the above algorithm is trivial.  If $f$ is not a linear or indicator function then step 5 in Alg.~\ref{alg:ARDr} will require an internal iterative algorithm.

\subsubsection{Sparsity properties of ARDr for orthogonal features}

\begin{figure}
\centering
\includegraphics[width=\textwidth]{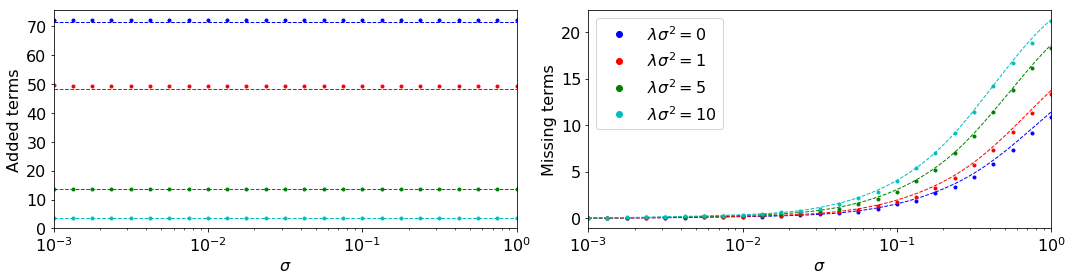}
\caption{Number of missing and added terms using ARDr using $250 x 250$ orthonormal $\btheta$ and holding $\lambda \sigma^2$ constant.  Dots indicate empirical average over 50 trials and dashed lined indicate values predicted by equations \eqref{eq:M_STSBL_FP_rate} and \eqref{eq:M_STSBL_FN_rate}. }
\label{fig:orthonormal_ARDr}
\end{figure}

The behavior of Alg. \ref{alg:ARDr} is complicated by the generality of functions $f$ and $g$.  In the simplest case we let $f$ be constant and $g$ be the linear function $g(\gamma_i) = \lambda \gamma_i$.  This is the formulation used in \cite{babacan2009bayesian} and a special case of using a Gamma distribution prior with shape parameter $k=1$ on $\gamma_i$.  The update step in line 5 of Alg.~\ref{alg:ARDr} gives $\gamma_i^{(k+1)} = |\xi_i^{(k+1)}|{c_i^{(k+1)}}^{-1/2}$ as in the unregularized case.  For a fixed point of Alg.~\ref{alg:ARDr} we have,
\begin{equation}
\begin{aligned}
c_i^* &= \bTheta_i^T \left( \sigma^2\bI + \bTheta \bGamma^*\bTheta^T \right)^{-1}\bTheta_i + \frac{\partial g}{\partial \gamma_i}\\
&= \dfrac{1}{\sigma^2\rho_i^{-1} + \frac{|\xi_i^*|}{\sqrt{c_i^*}}} + \lambda \\
0 &= \rho_i^{-1}\sigma^2 {c_i^*}^\frac{3}{2} + |\xi_i^*| {c_i^*} - \left( \lambda\rho_i^{-1} \sigma^2 +1 \right)  {c_i^*}^\frac{1}{2} -\lambda |\xi_i^*|
\end{aligned} \label{eq:ARDr_root_c*}
\end{equation}
The KKT conditions for the $\xi_i$ update are unchanged from the unregularized case and are given by,
\begin{equation}
0 \in \rho_i \xi_i^* - \rho_i \xi_i + \btheta^T_i\bnu + \sigma^2 \sqrt{c_i^*} sgn(\xi_i^*).
\label{eq:ARDr_KKT}
\end{equation}
If $\xi_i^* = 0$ then Eq.~\eqref{eq:ARDr_root_c*} tells us $\sqrt{c_i^*} = \sqrt{ \rho_i \sigma^{-2} + \lambda}$ and therefore,
\begin{equation}
\xi_i^* = 0  \Rightarrow \xi_i - \rho_i^{-1} \bTheta_i^T\bnu \in \left[ -  \sqrt{ \rho_i^{-1} \sigma^{2} + \lambda\rho_i^{-2} \sigma^{4}}, \sqrt{ \rho_i^{-1} \sigma^{2} + \lambda\rho_i^{-2} \sigma^{4}} \right] \label{eq:ARDr_xi0_cond}
\end{equation}
The converse of Eq. \eqref{eq:ARDr_xi0_cond} is shown in Appendix B.  From this equivalence it follows that the false positive and negative rates for Alg.~\ref{alg:ARDr} are given by,
\begin{equation}
FP_{r}(\lambda) = p(\xi_i^* \neq 0 | \xi_i = 0) = 1 - \text{erf}\left(\sqrt{  \frac{1 + \lambda \rho_i^{-1}\sigma^2}{2} }  \right)
\label{eq:ARDr_FP_rate}
\end{equation}
\vspace{-5mm} 
\begin{equation}
FN_{r}(\lambda) = \frac{1}{2} \left(  
\text{erf} \left( \frac{ \xi_i + \sqrt{ \rho_i^{-1} \sigma^{2} + \lambda\rho_i^{-2} \sigma^{4}}}{\sigma\sqrt{2\rho_i^{-1}}} \right) - 
\text{erf} \left( \frac{ \xi_i - \sqrt{ \rho_i^{-1} \sigma^{2} + \lambda\rho_i^{-2} \sigma^{4}}}{\sigma\sqrt{2\rho_i^{-1}}} \right) \right).
\label{eq:ARDr_FN_rate}
\end{equation}
\begin{figure}
\centering
\includegraphics[width=\textwidth]{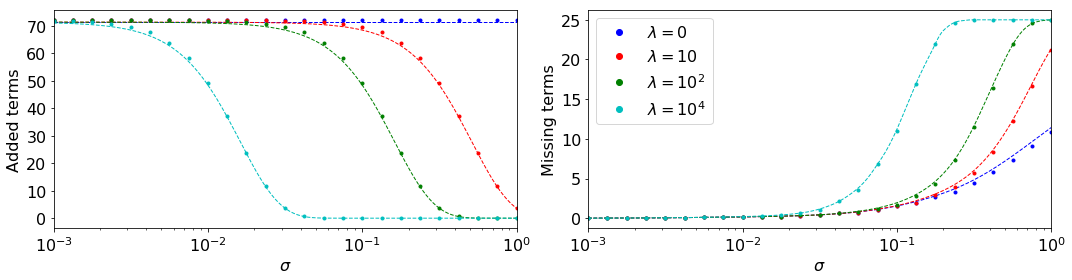}
\caption{Number of missing and added terms using ARDr using $250 x 250$ orthonormal $\bTheta$.  Dots indicate empirical average over 50 trials and dashed lined indicate values predicted by equations \eqref{eq:M_STSBL_FP_rate} and \eqref{eq:M_STSBL_FN_rate}. }
\label{fig:orthonormal_ARDr_2}
\end{figure}
These rates are verified empirically by testing 50 trials using 250x250 $\bTheta$ with orthogonal columns and random $\rho_i$ as in Sec. \ref{subsec:ARDvi}.  Results are shown in Fig.~\ref{fig:orthonormal_ARDr_2}.  Note that for fixed $\lambda > 0$ the false negative rate does indeed approach zero as $\sigma \to 0$, however, the false positive rate increases.  This indicates that a linear model with smaller error requires higher regularization to achieve a sparse solution.  For $\lambda \sigma^2$ held fixed as $\sigma$ varies, the false negative rate still approaches zero and the false positive rate is constant.  This latter case is shown in Fig.~\ref{fig:orthonormal_ARDr}.

Figure \ref{fig:orthonormal_ARDr} shows a similar convergence pattern to what we observed for ARDvi in Fig.~\ref{fig:orthonormal_variance_inflation}.  The number of added terms (false positives) remains constant as $\sigma \to 0$ for any fixed regularization parameter $\lambda$.  However, we note again that for sufficiently large $\lambda$ the fixed false positive rate may be made arbitrarily small.  In the following section we will construct thresholding methods including one for which the false positive and negative rates converge to zero as $\sigma \to 0$.

\section{Thresholding Based Methods}\label{sec:thresholding_methods}

As we have shown, automatic relevance determination will not realize the correct non-zero coefficients in a general sparse regression problem, but it will converge in any norm \cite{yuan2019machine}.  Therefore, applying an arbitrarily small threshold on $|\xi_i|$ will ensure selection of the correct nonzero coefficients in the limit of low noise.  In this section we discuss methods for thresholding the output from Alg.~\ref{alg:ARD} using the mean magnitude of coefficients $|\xi_i|$ or based on the posterior distribution of $\xi_i$.

\subsection{Magnitude Based Thresholding}

Sequential thresholding based on the magnitude of coefficients has been used extensively in regression \cite{Brunton2016, blumensath2009iterative} and also in conjunction with automatic relevance determination methods for identifying nonlinear dynamical systems with uncertainty quantification in \cite{zhang2018robust}.  Here we consider the method initially proposed in \cite{zhang2018robust}, called threshold sparse Bayesian regression (TSBR).  To distinguish from other thresholding methods we use the term magnitude sequential threshold sparse Bayesian learning (M-STSBL).  Magnitude based thresholding assumes that coefficients learned in the ARD algorithm with sufficiently small magnitude, $|\xi_j| < \tau$ are irrelevant and may be treated as zero.

\begin{algorithm}[H]
\begin{algorithmic}[1]
\STATE \vspace{1 mm} $\bgamma = \text{ARD}(\bTheta, \by, \sigma^2 )$
\STATE \vspace{1 mm} $\bxi = \sigma^{-2} \bSigma_\xi \bTheta^T \by$
\STATE \vspace{1 mm} $\mathcal{G} = \{i : |\xi_i| \geq \tau \}$
\STATE \vspace{1 mm} $\gamma_{\mathcal{G}^c}= 0$
\STATE \vspace{1 mm} if $\mathcal{G}^c \neq \emptyset $:   $\gamma_\mathcal{G} = \text{M-STSBL}(\bTheta_\mathcal{G} , \by, \sigma^2, \tau)$
\STATE \vspace{1 mm} return $\bgamma$
\end{algorithmic}
\caption{M-STSBL($\bTheta , \by, \sigma^2, \tau$)}
\label{alg:M-STSBL}
\end{algorithm}

The sequential hard-thresholding method for automatic relevance determination is implemented in Alg. \ref{alg:M-STSBL}.  Non-zero terms are indexed by $\mathcal{G}$ whose complement $\mathcal{G}^c$ tracks terms removed from the model.  At each iteration the algorithm either recursively calls itself with fewer features or terminates if all features are kept non-zero.

\subsubsection{Sparsity properties of M-STSBL for orthogonal features}

\begin{figure}
\centering
\includegraphics[width=\textwidth]{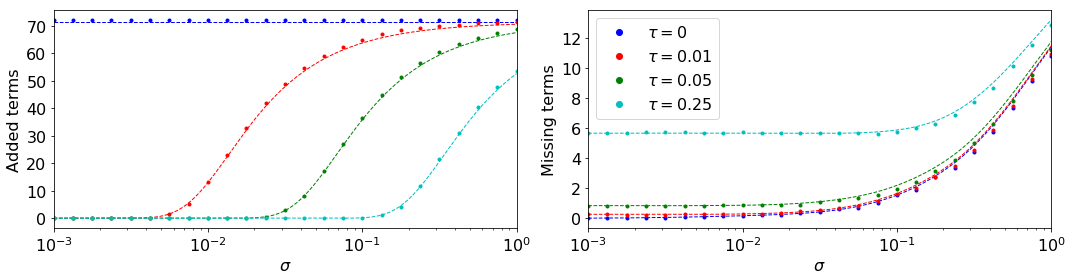}
\caption{Number of missing and added terms using M-STSBL using $250 x 250$ orthonormal $\bTheta$.  Dots indicate empirical average over 50 trials and dashed lined indicate values predicted by equations \eqref{eq:M_STSBL_FP_rate} and \eqref{eq:M_STSBL_FN_rate}.}
\label{fig:orthonormal_threshodling}
\end{figure}

We consider the number of errors using Alg.~\ref{alg:M-STSBL} in a similar context to the analysis of the variance inflation and regularized method.  First consider the likelihood of a false non-zero term.  Recall from the previous section that the KKT conditions for a fixed point of Alg. \ref{alg:ARD} imply,
\begin{equation}
\begin{aligned}
0 &\in  \xi^*_i - \xi_i + \rho_i^{-1}\bTheta_i^T\bnu+ \rho_i^{-1}\sigma^2 \sqrt{c_i^*} \partial \|\xi_i^*\|_1 \\
& = \xi^*_i - \xi_i + \rho_i^{-1}\bTheta_i^T\bnu+ \frac{1}{2} \left( \sqrt{{\xi_i^*}^2 + 4\rho_i^{-1}\sigma^2} - |\xi_i^*| \right) sgn ( \xi_i^* ),
\end{aligned}
\end{equation}
We can rewrite this as,
\begin{equation}
\phi_{\sigma, \rho} (\xi_i^*) = \xi_i-\rho_i^{-1} \bTheta_i^T\bnu \sim \mathcal{N}(0, \rho_i^{-1}\sigma^2)
\end{equation}
where,
\begin{equation}
\begin{aligned}
\phi_{\sigma, \rho} (\xi_i^*) &= \frac{\xi_i^*}{2} + \frac{1}{2} \sqrt{\xi_i^2 + 4\rho_i^{-1}\sigma^2  } sgn(\xi_i^*) \\
&= \frac{\xi_i^*}{2} \left( 1 + \sqrt{1 + 4\rho_i^{-1}\sigma^2 {\xi_i^*}^{-2} } \right), \text{   for   } \xi_i^* \neq 0,
\end{aligned}
\end{equation}
is invertible on $\bbR \setminus \{0\}$ and strictly increasing.  Therefore,
\begin{equation}
\left|\xi_i - \rho_i^{-1}\bTheta_i^T\bnu\right| > \phi_{\sigma,\rho}(\tau) \Leftrightarrow |\xi_i^*| > \tau.
\label{eq:phi_transform}
\end{equation}
This gives the likelihood of a false non-zero coefficient as,
\begin{equation}
FP_M(\tau) = p(\xi_i^* \neq 0 | \xi_i = 0) = 1 - \text{erf} \left( \frac{\phi_{\sigma,\rho} (\tau)}{\sigma\sqrt{2\rho_i^{-1}}}  \right),
\label{eq:M_STSBL_FP_rate}
\end{equation}
and the likelihood for a false zero coefficient as,
\begin{equation}
FN_M(\tau) = p(\xi_i^* = 0 | \xi_i \neq 0) = \frac{1}{2} \left(  \text{erf} \left( \frac{\xi_i +  \phi_{\sigma,\rho}(\tau) }{\sigma\sqrt{2\rho_i^{-1}}}  \right) -  \text{erf} \left(  \frac{ \xi_i - \phi_{\sigma,\rho}(\tau)  }{\sigma\sqrt{2\rho_i^{-1}}}  \right)   \right).
\label{eq:M_STSBL_FN_rate}
\end{equation}
Equations \eqref{eq:M_STSBL_FN_rate} and \eqref{eq:M_STSBL_FP_rate} are verified empirically by testing on 50 trials over a 250x250 $\bTheta$ using the same experimental design as in Sec. \ref{subsec:ARDvi}.  Results shown in Fig.~\ref{fig:orthonormal_threshodling}.  In contrast to regularization based approaches, we now have the desirable condition where the number of false positive terms each goes to zero as $\sigma \to 0$.  However, the number false negatives now only shrinks to a fixed positive number - a consequence of using a hard threshold.  This motivates alternative criteria for thresholding.  In the next section, we will discuss thresholding based not strictly on magnitude but on the marginal posterior likelihood that a coefficient is zero.

\subsection{Likelihood Based Thresholding}

While Alg.~\ref{alg:M-STSBL} was shown to be effective in \cite{zhang2018robust} it is not independent from the units of measurement used for each feature and is not practical in the case where some true coefficients are small.  An alternative means of thresholding is to do so based on the marginal likelihood of a coefficient being zero.  The marginal posterior distribution of $\xi_i$ is given by,
\begin{equation}
p(\xi_i) \sim \mathcal{N} (\mu_{\xi,i}, \Sigma_{\xi, ii}),
\end{equation}
where $\mu_{\xi,i}, \Sigma_{\xi, ii}$ are given by Eq.\eqref{eq:xi_posterior} and the marginal likelihood that $\xi_i=0$ is,
\begin{equation}
p(\xi_i=0  ) = \mathcal{N} (0 \,| \, \mu_{\xi,i}, \Sigma_{\xi, ii}) =  \frac{1}{\sqrt{2\pi \Sigma_{\xi, ii}}} e^{-\frac{1}{2}\mu_{\xi,i}^2\Sigma_{\xi,ii}^{-1}}.
\end{equation}
We can construct a sequential thresholding algorithm shown by Alg.~\ref{alg:L-STSBL} by removing terms whose marginal likelihood evaluated at zero is sufficiently large.  The remaining subset of features in then passed recursively to the same procedure until convergence, marked by no change in the number of features.  This process is described by Alg.~\ref{alg:L-STSBL} where parameter $\tau$ is the marginal likelihood at zero above which features are removed.

\begin{algorithm}
\begin{algorithmic}[1]
\STATE \vspace{1 mm} $\bgamma = \text{ARD}(\bTheta, \by, \sigma^2 )$
\STATE \vspace{1 mm} $\bxi = \sigma^{-2} \bSigma_\xi \bTheta^T \by$
\STATE \vspace{1 mm} $\mathcal{G} = \left\{i : (2\pi\Sigma_{\xi,ii})^{-1/2}exp(-\frac{1}{2}\xi_i^2\Sigma_{\xi,ii}^{-1}) \leq \tau \right\}$
\STATE \vspace{1 mm} $\bgamma_{\mathcal{G}^{c} }= 0$
\STATE \vspace{1 mm} if $\mathcal{G}^c \neq \emptyset $:   $\gamma_\mathcal{G} = \text{L-STSBL}(\bTheta_\mathcal{G} , \by, \sigma^2, \tau)$
\STATE \vspace{1 mm} return $\bgamma$
\end{algorithmic}
\caption{L-STSBL($\bTheta , \by, \sigma^2, \tau$)}
\label{alg:L-STSBL}
\end{algorithm}

\subsubsection{Sparsity properties of L-STSBL for orthogonal features} \label{subsec:L_STSBL_sparsity}

We again consider the case where $\bTheta^T\bTheta = diag(\brho)$.  Let,
\begin{equation}
h_L (\xi_i , \Sigma_{\xi,ii}) = (2\pi\Sigma_{\xi,ii})^{-1/2}exp\left(-\frac{1}{2}\xi_i^2\Sigma_{\xi,ii}^{-1}\right), \label{eq:h_l}
\end{equation}
so that the thresholding criteria is $h_L (\xi_i , \Sigma_{\xi,ii}) > \tau$.
In Alg. \ref{alg:ARD} $\boldsymbol{\eta^*} = 2\sigma^2 \sqrt{\bc^{*}} = 2\sigma^2 \bxi^* {\bgamma^*}^{-1}$ so $\bxi^* = \sigma^{-2} \bSigma_\xi \bTheta^T \by$ is the mean posterior estimate.  The orthogonality of the columnsof $\bTheta$ allows us to express the marginal posterior variance $\Sigma_{\xi, ii}$ as a function of $|\xi_i|$.  Letting $\bSigma_{\xi}^*$ be the covariance from a fixed point of Alg. \ref{alg:ARD} we have,
\begin{equation}
\begin{aligned}
{\Sigma_{\xi, ii}^*} &= \left( \sigma^{-2} \bTheta^T\bTheta + {\bGamma^*}^{-1} \right)^{-1}_{ii} \\
&= \left( \sigma^{-2} \diag(\brho) + {\bGamma^*}^{-1} \right)^{-1}_{ii} \\
&= \left(\frac{\rho_i}{\sigma^2} + {\gamma_i^*}^{-1}\right)^{-1}   \\
{\Sigma_{\xi}^*}^{-1}_{ii} &= \frac{\rho_i}{\sigma^2} + \frac{\sqrt{c_i^*}}{|\xi_i^*|}   \\
&= \frac{\rho_i}{\sigma^2} + \frac{\sqrt{1+4\rho_i^{-1}\sigma^2|\xi_i^*|^{-2}}-1}{2\rho_i^{-1}\sigma^2} \\
&= \frac{\sqrt{1+4\rho_i^{-1}\sigma^2|\xi_i^*|^{-2}}+1}{2\rho_i^{-1}\sigma^2} .
\end{aligned}
\label{eq:diagonal_post_var}
\end{equation} 
This allows us to express,
\begin{equation}
h_L (\xi_i^* , \Sigma_{\xi,ii}^*) = \tilde{h}_{L,\rho,\sigma} (|\xi_i^*|) = (2\pi)^{-1/2} \sqrt{{\Sigma_{\xi,ii}^*}^{-1}(|\xi_i|)} \, exp\left(-\frac{1}{2}|\xi_i^*|^2 {\Sigma_{\xi,ii}^*}^{-1} (|\xi_i|)\right),
\end{equation}
where ${\Sigma_{\xi,ii}^*}^{-1}(|\xi_i^*|)$ and $|\xi_i^*|^2 {\Sigma_{\xi,ii}^*}^{-1}(|\xi_i^*|)$ are strictly decreasing and increasing functions of $|\xi_i^*|$, respectively.  It follows that $\tilde{h}_{L,\rho,\sigma}$ is strictly decreasing and therefore invertible with $\tilde{h}_{L,\rho,\sigma}^{-1}$ easily computed by bisection.  For $\tau > 0$ there is some $\tilde{h}_{L,\rho,\sigma}^{-1}(\tau)$ such that,
\begin{equation}
|\xi_i^*| > \tilde{h}_{L,\rho,\sigma}^{-1}(\tau) \Leftrightarrow \tilde{h}_{L,\rho,\sigma}(\xi_i^*) \leq \tau,
\end{equation}
and recalling Eq.~\eqref{eq:phi_transform},
\begin{equation}
|\xi_i - \rho_i^{-1}\bTheta_i^T\bnu| > \phi_{\sigma, \rho}\left(\tilde{h}_{L,\rho,\sigma}^{-1}(\tau)\right) \Leftrightarrow \tilde{h}_{L,\rho,\sigma}(\xi_i^*) \leq \tau.
\end{equation}
This gives,
\begin{equation}
FP_L(\tau) = p(\xi_i^* \neq 0 | \xi_i = 0) = 1 - \text{erf} \left( \frac{\phi_{\sigma, \rho}\left(\tilde{h}_{L,\rho,\sigma}^{-1}(\tau)\right)}{\sigma\sqrt{2\rho_i^{-1}}}  \right),
\label{eq:L_STSBL_FP_rate}
\end{equation}
and,
\begin{equation}
FN_L(\tau) = \frac{1}{2} \left(  \text{erf} \left( \frac{\xi_i +  \phi_{\sigma, \rho}\left(\tilde{h}_{L,\rho,\sigma}^{-1}(\tau)\right) }{\sigma\sqrt{2\rho_i^{-1}}}  \right) -  \text{erf} \left(  \frac{ \xi_i - \phi_{\sigma, \rho}\left(\tilde{h}_{L,\rho,\sigma}^{-1}(\tau)\right)  }{\sigma\sqrt{2\rho_i^{-1}}}  \right)   \right).
\label{eq:L_STSBL_FN_rate}
\end{equation}

\begin{figure}
\centering
\includegraphics[width=\textwidth]{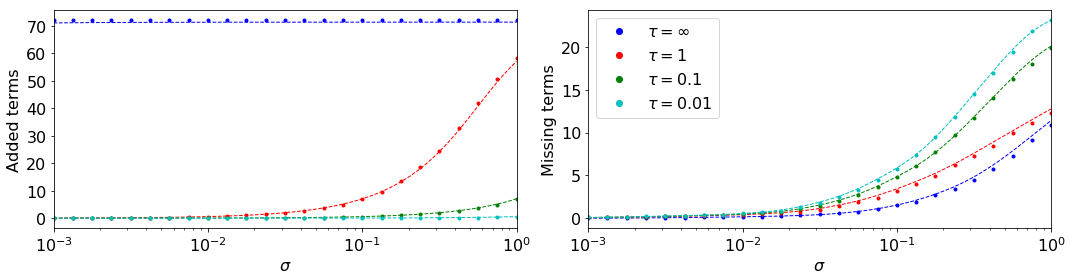}
\caption{Number of missing and added terms using L-STSBL using $250 x 250$ orthonormal $\bTheta$.  Dots indicate points where data was collected as the empirical average over 50 trials.}
\label{fig:orthonormal_L_STSBL}
\end{figure}

Equations \eqref{eq:L_STSBL_FN_rate} and \eqref{eq:L_STSBL_FP_rate} are verified empirically using the same experimental setup as in previous sections.  Results shown in Fig.~\ref{fig:orthonormal_L_STSBL}.  Similar to M-STSBL, solutions of L-STSBL converge towards the correct sparsity pattern as $\sigma \to 0$.  However, Fig.~\ref{fig:orthonormal_L_STSBL} indicates highly favorable results in the number of missing terms.  Eq. \eqref{eq:diagonal_post_var} indicates that for $\sigma \ll 1$ the marginal variance $\Sigma_{\xi,ii} \sim \mathcal{O}(\sigma^2)$.  As a consequence, the exponential in Eq. \eqref{eq:h_l} becomes very small and the algorithms is much more conservative about pruning terms.

\subsection{Thresholding via Sparse Prior on $\boldsymbol{\xi}$}

Algorithm \ref{alg:L-STSBL} performs thresholding based on the marginal likelihood of a given coefficient being zero without consideration for the likelihood of the coefficient prior to applying a threshold. We now propose a thresholding method which includes the latter.  We consider a prior on $\bxi$ which varies from Eq. \eqref{eq:xi_prior} only where $\|\bxi\|_0 < 0$ and use MAP estimates of $\bxi$ to prune terms.  Consider the same model described in Sec. \ref{sec:setup} but with the following prior on $\xi_i$,
\begin{equation}
p(\xi_i) = \cN (\xi_i | 0, \gamma_i)\, e^{\tau \delta_{\xi_i, 0}}.
\label{eq:sparse_xi_prior}
\end{equation}
Note that this is equivalent to \eqref{eq:xi_prior} almost everywhere so the integral in \eqref{eq:gamma_posterior} is not affected.  The posterior for $\xi$ under assumption \eqref{eq:sparse_xi_prior} is then,
\begin{equation}
    p(\bxi | \mathcal{D}, \tau) \propto \dfrac{1}{(2\pi)^{d/2}|\bSigma_\xi|} \, e^{-\frac{1}{2}(\bxi-\bmu_\xi)^T\bSigma_\xi^{-1}(\bxi-\bmu_\xi) - \tau\|\bxi\|_0}
    \label{eq:sparse_xi_posterior}
\end{equation}
with $\bmu_{\xi}, \bSigma_\xi$ defined as in Eq. \eqref{eq:xi_posterior}.  When $\tau=0$ this reduces to the standard ARD posterior but for $\tau>0$ the likelihood shrinks exponentially in the number of nonzero terms.  Since the two posteriors differ only on a set of measure zero, the inclusion of $exp(\tau \delta_{\xi_i,0})$ in Eq. \eqref{eq:sparse_xi_prior} only affects the solution if we use the MAP estimate of $\bxi$ as a means to select active terms.  Doing so induces a thresholding operation to find $\bxi_{MAP}$.  

For a group $S = \{s_1, s_2, \hdots, s_q\} \subseteq \{1, \hdots , d\}$ where $\bmu_{\xi, s_i} \neq 0$ let $\bxi_{-S} = \bmu_\xi - \sum_q\mu_{\xi, s_i}\be_{s_i}$ where $\be_i$ is the unit vector in the $\text{i}^\text{th}$ coordinate.  The likelihood of the thresholded vector $\bxi_{-S}$ is given by,
\begin{equation}
\begin{aligned}
p(\bxi_{-S} | \mathcal{D}, \tau) &= C \, exp\left(-\frac{1}{2}(\bxi_{-S}-\bmu_\xi)^T\bSigma_\xi^{-1}(\bxi_{-S}-\bmu_\xi) - \tau\|\bxi_{-S}\|_0\right) \\
&= C \, exp\left(-\frac{1}{2}\left(\sum_q \mu_{\xi, s_i}\be_{s_i}\right)^T\bSigma_\xi^{-1}\left(\sum_q \mu_{\xi, s_i}\be_{s_i}\right) - \tau(\|\bmu_\xi\|_0 - q )\right) \\
&= p(\bmu_\xi | \mathcal{D}, \tau) exp\left( -\frac{1}{2} \bmu_{\xi,S}^T\bSigma_{\xi,S}^{-1}\bmu_{\xi,S} + q\tau  \right),
\end{aligned} \label{eq:multi_thresh_likelihood}
\end{equation}
where $\bSigma_{\xi,S}^{-1}$ is the square sub-matrix of $\bSigma_{\xi}^{-1}$ formed by the rows and columns indexed by $S$.  Then,
\begin{equation}
p(\bxi_{-S} | \mathcal{D}, \tau) > p(\bmu_\xi | \mathcal{D}, \tau) \text{   if   } \frac{1}{2} \bmu_{\xi,S}^T\bSigma_{\xi,S}^{-1}\bmu_{\xi,S} < q\tau, \label{eq:multi_thresh_criteria}
\end{equation}
and the MAP estimate of $\xi$ is given by,
\begin{equation}
\bxi_{MAP} = \bxi_{-S} \text{   where   } S = \underset{S \in \mathcal{P}([d])}{arg\,max} \,\, p(\bxi_{-S} | \mathcal{D}, \tau) \label{eq:sparse_xi_MAP}
\end{equation}
Equation \eqref{eq:sparse_xi_MAP} is combinatorially hard so we approximate it in a manner that makes solution tractable.  Most simply we can treat the precision matrix $\bSigma_\xi^{-1}$ as diagonal so that decisions with regards to each variable are decoupled.  Alternatively, we can use a greedy algorithm to construct the $S$ maximizing Eq. \eqref{eq:sparse_xi_MAP}.  In this case we iteratively add to $S$ the most likely additional term until no term increases the likelihood.  The algorithm may be further refined as a forward-backward greedy algorithm.  Here we restrict our attention to the diagonal approximation of the posterior covariance.  This gives the simple threshold,
\begin{equation}
\xi_i = 0 \text{ if } \frac{1}{2} \mu_{\xi, i}^2 \Sigma_{\xi, ii}^{-1} < \tau, \label{eq:single_thresh_criteria}
\end{equation}
which is implemented in Alg.~\ref{alg:MAP-STSBL}.  The same pruning technique has also been used for connections in Bayesian neural networks, using the variational approximation of the posterior \cite{graves2011practical}.  We call this technique maximum a-posteriori sequential threshold sparse Bayesian learning (MAP-STSBL). 
\begin{algorithm}
\begin{algorithmic}[1]
\STATE \vspace{1 mm} $\bgamma = \text{ARD}(\bTheta, \by, \sigma^2 )$
\STATE \vspace{1 mm} $S \approx \underset{S \in \mathcal{P}([d])}{arg\,max} \,\, p(\bxi_{-S} | \mathcal{D}, \tau)$
\STATE \vspace{1 mm} $\bgamma_{-S}= 0$
\STATE \vspace{1 mm} if $|S| \neq 0 $:   $\gamma_S = \text{MAP-STSBL}(\bTheta_S , \by, \sigma^2, \tau)$
\STATE \vspace{1 mm} return $\bgamma$
\end{algorithmic}
\caption{MAP-STSBL($\bTheta , \by, \sigma^2, \tau$)}
\label{alg:MAP-STSBL}
\end{algorithm}

\subsection{Sparsity properties of MAP-STSBL for orthogonal features}

\begin{figure}
\centering
\includegraphics[width=\textwidth]{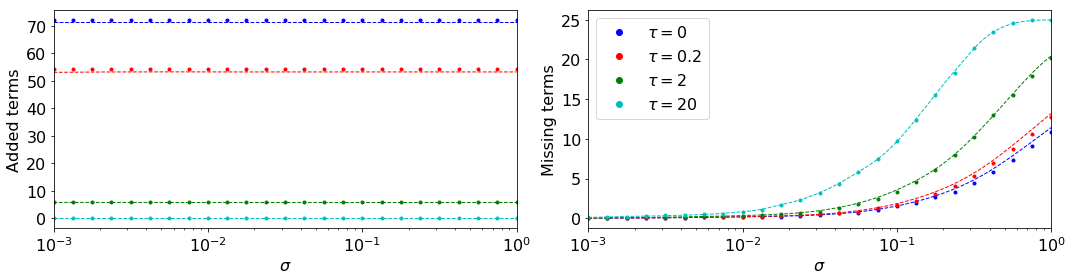}
\caption{ Number of missing and added terms using MAP-STSBL using $250 x 250$ orthonormal $\bTheta$.  Dots indicate points where data was collected as the empirical average over 50 trials. }
\label{fig:orthonormal_MAP_STSBL}
\end{figure}

In the case of orthogonal columns of $\bTheta$ we can use the same simplification as in Sec. \ref{subsec:L_STSBL_sparsity} to simplify the thresholding criteria in Eq. \eqref{eq:single_thresh_criteria} to,
\begin{equation}
h_{MAP}(\xi_i^*, {\Sigma_{\xi,ii}^*}) = \frac{1}{2} |\xi_i^*|^2 {\Sigma_{\xi,ii}^*}^{-1} = \frac{|\xi_i^*|^2 \left(\sqrt{1+4\rho_i^{-1}\sigma^2|\xi_i^*|^{-2}}+1\right)}{4\rho_i^{-1}\sigma^2} = \tilde{h}_{MAP}(|\xi_i^*|)
\end{equation}
which has inverse given by,
\begin{equation}
\tilde{h}_{MAP}^{-1}(\tau) = 2\sqrt{\frac{\rho_i^{-1}\sigma^2\tau^2}{1+2\tau}}
\end{equation}
Then for $\tau > 0$ there is $\tilde{h}_{MAP}^{-1}(\tau)$ such that,
\begin{equation}
|\xi_i^*| > \tilde{h}_{MAP}^{-1}(\tau) \Leftrightarrow \tilde{h}_{MAP}(\xi_i^*) \geq \tau,
\end{equation}
and,
\begin{equation}
|\xi_i - \rho_i^{-1}\bTheta_i^T\bnu| > \phi_{\sigma, \rho}\left(\tilde{h}_{MAP}^{-1}(\tau)\right) = \sigma \sqrt{\rho_i^{-1} (2\tau+1)} \Leftrightarrow \tilde{h}_{MAP}(\xi_i^*) \leq \tau.
\end{equation}
This gives,
\begin{equation}
FP_{MAP}(\tau) = p(\xi_i^* \neq 0 | \xi_i = 0) = 1 - \text{erf} \left( \sqrt{\frac{2\tau+1}{2}} \right),
\label{eq:MAP_STSBL_FP_rate}
\end{equation}
and,
\begin{equation}
FN_{MAP}(\tau) = \frac{1}{2} \left(  \text{erf} \left( \frac{\xi_i +  \sigma \sqrt{\rho_i^{-1} (2\tau+1)} }{\sigma\sqrt{2\rho_i^{-1}}}  \right) -  \text{erf} \left(  \frac{ \xi_i - \sigma \sqrt{\rho_i^{-1} (2\tau+1)}  }{\sigma\sqrt{2\rho_i^{-1}}}  \right)   \right).
\label{eq:MAP_STSBL_FN_rate}
\end{equation}
Equations \eqref{eq:MAP_STSBL_FP_rate} and \eqref{eq:MAP_STSBL_FN_rate} are verified empirically in Fig. \ref{fig:orthonormal_MAP_STSBL}.  The results are very similar to those for ARDvi.  Indeed, equations \eqref{eq:MAP_STSBL_FP_rate} and \eqref{eq:MAP_STSBL_FN_rate} show that for orthogonal features there is a transformation $\alpha \to 2\tau+1$ under which ARDvi and MAP-STSBL realize the same sparsity pattern. We will show empirically in a subsequent section that this is not true in the case where columns of $\bTheta$ are not orthogonal.

\section{Comparison}\label{sec:comparison}

\begin{figure}
\centering
\includegraphics[width=0.5\textwidth]{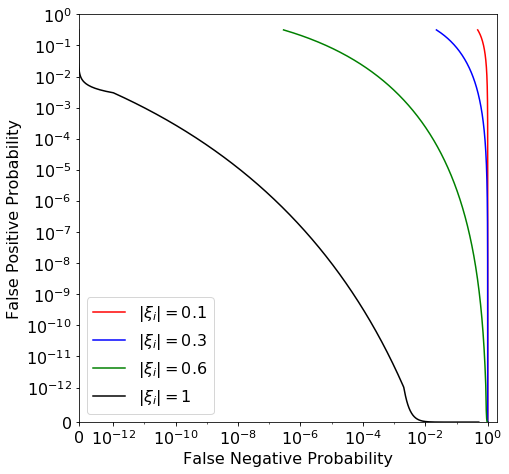}
\caption{The FP/FN curve for orthogonal matrices $\bTheta^T\bTheta = \bI$.}
\label{fig:FP_FN_curve}
\end{figure}

\begin{figure}
\centering
\includegraphics[width=\textwidth]{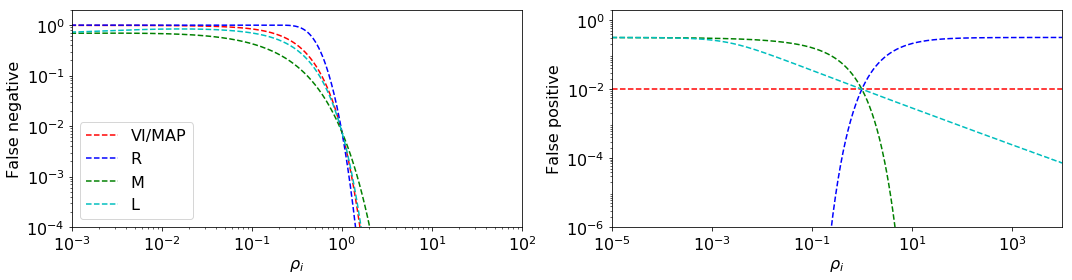}
\caption{FP/FN rates for a term with $\|\bTheta_i\|^2 = \rho_i$.  Parameters for each method have been selected such that $\text{FP}|_{\rho_i = 1} = \text{FN}|_{\rho_i = 1}$.}
\label{fig:FP_FN_rho}
\end{figure}

The false positive and negative likelihoods for $\xi_i$ each of the methods discussed in Sections \ref{sec:regularization_methods} and \ref{sec:thresholding_methods} are summarized by,
\begin{equation}
\begin{aligned}
FP_{\bullet}(\xi_i ; \omega) &= 1 - \text{erf} \left( \frac{\psi_{\bullet} (\omega)}{\sigma \sqrt{2 \rho^{-1}}}  \right) \\
FN_{\bullet}(\xi_i ; \omega) &= \frac{1}{2} \left(
\text{erf} \left( \frac{\xi_i + \sigma \psi_{\bullet} (\omega)}{\sigma \sqrt{2 \rho^{-1}}}  \right) -
\text{erf} \left( \frac{\xi_i - \sigma \psi_{\bullet} (\omega)}{\sigma \sqrt{2 \rho^{-1}}}  \right) \right)
\end{aligned}
\end{equation}
where $\bullet$ refers to the method, $\omega$ to the input ($\alpha$, $\lambda$, or $\tau$) and,
\begin{equation}
\begin{aligned}
\psi_{ARDvi}(\alpha) &= \sigma \sqrt{\alpha \rho_i^{-1}} \\
\psi_{ARDr}(\lambda) &= \sqrt{\rho_i^{-1}\sigma^2 + \lambda\rho_i^{-2}\sigma^4}  \\
\psi_{M-STSBL}(\tau) &= \phi_{\sigma, \rho}\left(\tau \right)\\
\psi_{L-STSBL}(\tau) &= \phi_{\sigma, \rho}\left(\tilde{h}_{L,\rho,\sigma}^{-1}(\tau)\right) \\
\psi_{MAP-STSBL}(\tau) &= \sigma \sqrt{(2\tau+1) \rho_i^{-1}}.
\end{aligned}
\end{equation}
Note that if $\rho_i = \rho_j$ for all $i$, $j$ then the false positive and negative rates are all equivalent under transformations of the parameters used for each method.
Curves $(FP_{\bullet}(\xi_i ; \omega), FN_{\bullet}(\xi_i ; \omega))$ parameterized by $\psi (\omega )$ are shown in Fig. \ref{fig:FP_FN_curve} for several values of $\xi_i$.
If $\rho_i$ are unequal then the specific parameter pair that will yield similar results for one column given two different algorithms will not hold for another column.  Hence, the methods differ in how they scale with $\rho_i$.  The exception is the pair ARDvi and MAP-STSBL which have the same false positive and negative rates for any $\rho_i$ under the transformation $\alpha = 2\tau+1$.  
To visualize the dependence of each false positive and false negative rate on $\rho_i$ we find parameters $\omega_\bullet$ for each method such that the $FP_\bullet (\omega_\bullet ) = FN_\bullet (\omega_\bullet )$ when $\rho_i=1$ and plot the resulting rates over a range of $\rho_i$.  This is shown in Fig. \ref{fig:FP_FN_rho}.  
The false negative rate for each method decreases monotonically in $\rho_i$.  This is intuitive, since larger $\rho_i$ corresponds to that term having a larger effect on $\by$.  The false negative rate as a function of $\rho_i$ is constant for ARDvi and MAP-STSBL, decreasing for L-STSBL and M-STSBL and increasing for ARDr.  These trends are explained by the asymptotic behavior of $\psi_{\bullet}$ for large $\rho$.  We have,
\begin{equation}
\begin{aligned}
\psi_{vi/MAP}(\alpha) &\sim \mathcal{O}(\rho^{-1/2}) \\
\psi_{ARDr}(\lambda) &\sim \mathcal{O}(\rho^{-1}) \\
\psi_{M-STSBL}(\tau) &\sim \mathcal{O}(1) \\
\psi_{L-STSBL}(\tau) &\sim \mathcal{O}(\rho^q), \, q\in (-1/2,0) \\
\end{aligned}
\label{eq:large_rho}
\end{equation}
where the last statement is inferred from Fig. \ref{fig:FP_FN_rho}.  Since $\psi_{\bullet}$ is multiplied by $\rho^{1/2}$ in the expression for the false positive rate, which is a decreasing function of $\psi_{\bullet}$, the trends in Fig. \ref{fig:FP_FN_rho} follow from Eq.~\eqref{eq:large_rho}.  If we allow ourselves to equate $\rho_i$ with sample size, then M-STSBL and L-STSBL have the desireable property that the false positive rate is decreases.

For orthogonal covariates, ARDvi and MAP-STSBL have equivalent behavior with regards to expected sparsity.  However, they begin to yield different results in that case that columns of $\bTheta$ are correlated.  This may be the result of the MAP-threshold criteria no longer aligning with the increased sparsity due to inflated variance, but there is also a fundamental change in the thresholding algorithms which occurs when we move away from orthogonal covariates.
We have shown that when $\bTheta^T\bTheta$ is diagonal the sparsity of $\xi_i$ depends only on the the inner product of the error $\bnu$ with $\bTheta_i$.  Hence, the recursion defined in each thresholding algorithm terminates at a depth of one.  This is not true when columns are correlated.  For dense $\bTheta^T\bTheta$ the recursion limit is the number of columns, though the algorithm tends to terminate far earlier.

\begin{figure}
\centering
\includegraphics[width=\textwidth]{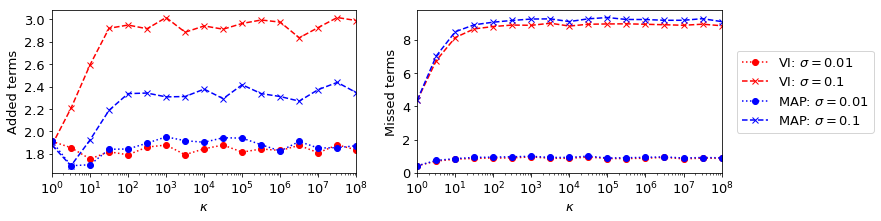}
\caption{FP/FN rates ARDvi and MAP-STSBL with two noise magnitudes as a function of $\kappa=\text{Cond}(\bTheta)$, the condition number of $\bTheta$.  Datapoints are averaged over 1000 trials using $\bTheta \in \bbR^{100\times 100}$ and singular values evenly spaced in $[\kappa^{-1}, 1]$.}
\label{fig:VI_MAP_k100}
\end{figure}

\begin{figure}
\centering
\includegraphics[width=0.5\textwidth]{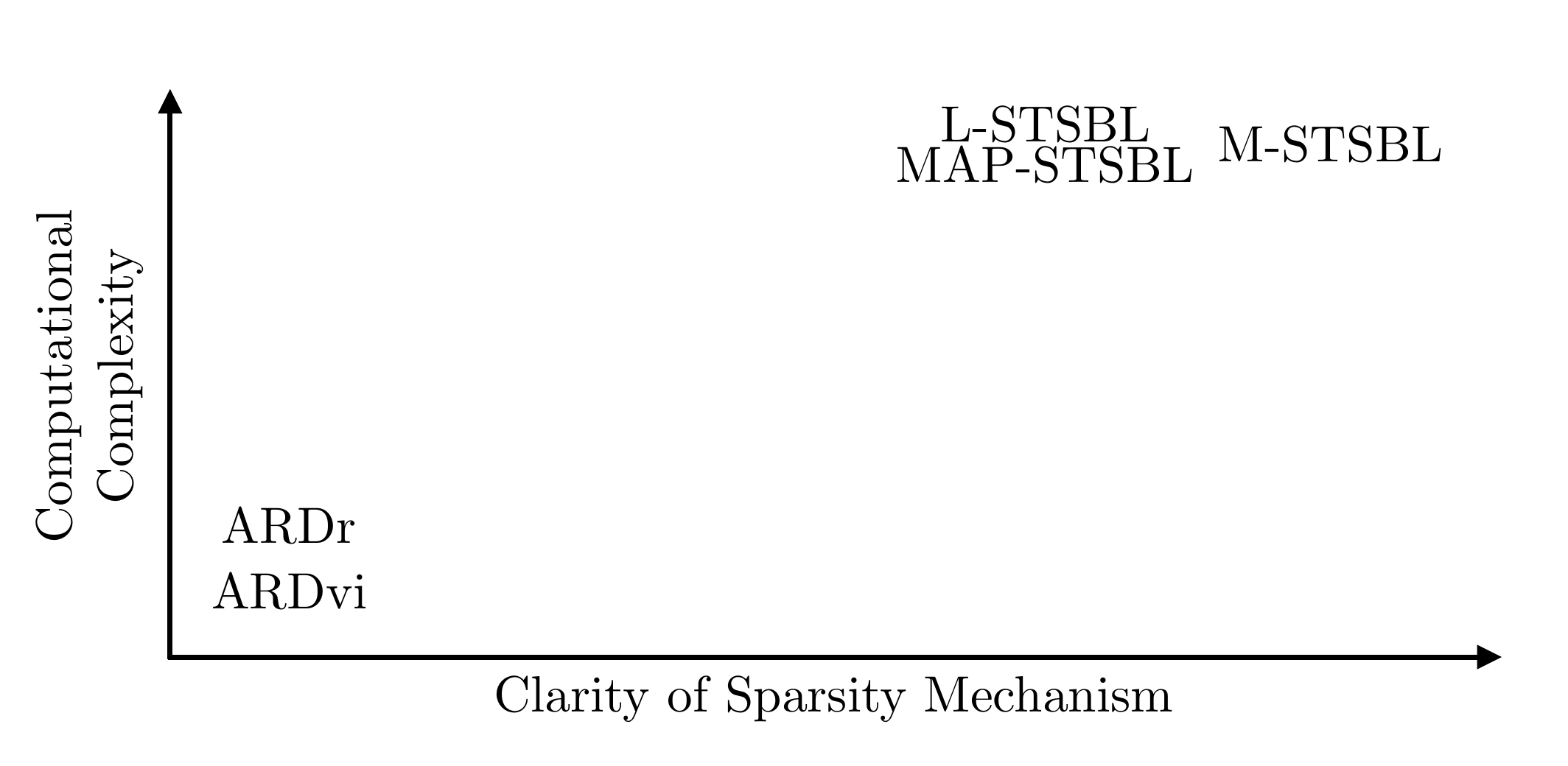}
\caption{Simple comparison of the relative merits of the five proposed methods.}
\label{fig:comparison_fig}
\end{figure}

An analytical comparison between the algorithms considered in the previous sections for non-orthogonal data is beyond the scope of this work.  However, there is one clear trade off between computational complexity and clarity of the algorithm's mechanism for inducing sparsity.  Thresholding algorithms offer clear criteria for setting additional terms to zero since we know the magnitude or likelihood at which a coefficient was pruned and at what step.  Regularization methods do not provide the same clarity but avoid the cost of increased computational time due to recursion.  In particular, for problems with many covariates, the depth limit in the thresholding algorithms is high.  We consider M-STSBL to be slightly more clear than MAP-STSBL and L-STSBL, since the thresholding parameter is a magnitude.  We initialize ARDr using ARD, so it is slightly more expensive than ARDvi.  This is summarized in Fig.~\ref{fig:comparison_fig}.  We will present several examples in the following section to compare the algorithms' performance on an empirical basis.

\section{Numerical experiments}\label{sec:results}

In this section we compare the performance of each of the methods considered in this work on several test problems.  These inlude a $250$ dimensional linear problem, function fitting in a Fourier basis, and system identification for the Lorenz 63, Lorenz 96, and Kuramoto Sivashinsky equations.  We test each of the proposed methods using a range of input parameters.  For the linear and function fitting examples we select optimal parameters for the regularizatio with the Akaike Information Criterion \cite{akaike1974new} with small sample size correction (AICc) \cite{cavanaugh1997unifying}, given by,  
\begin{equation}
AIC_c(\bgamma) = 2k - 2 \ln \left(p(\bgamma)\right) = 2k - \underset{\bxi}{min} \, \left(log |\bSigma_y| + \frac{1}{\sigma^2} \|\by - \bTheta \bxi \|_2^2 + \bxi^T\bGamma^{-1}\bxi  \right),
\label{eq:AICc}
\end{equation}
where $k = \|\bgamma\|_0+1$ is the number of terms fit by the model including error variance $\sigma^2$.  For consistency across methods, we do not consider regularization terms when evaluating the likelihood.  For examples of nonlinear system identification we found $AIC_c$ selected models with extraneous variables even when the true model was available.  This is perhaps due to the errors being non-Gaussian and correlated between observations, since numerical differentiation uses adjacent points.  We therefore select optimal regularization parameters for the system identification based on minimal mismatch in sparsity to the true solution.  This is not practical in an application setting but highlights differences between the algorithms presented in this work without the need for more robust model selection.

Algorithm~\ref{alg:ARDr} allows for substantial freedom in the choice of specific regularization functions $f$ and $g$.  For the purposes of comparing with other methods discussed in this work we restrict our attention to the case where $f$ is a constant and $g(\gamma_i) = \lambda \sigma^{-2} \, min\{\gamma_i, \eta\}$ which is constant for $\gamma_i > \eta$ and linear with positive slope $\lambda \sigma^{-2}$ for $\gamma_i \leq \eta$.  We search over parameter $\lambda$, keeping $\eta$ fixed at a value of $0.1$.  It is reasonable to assume that Alg.~\ref{alg:ARDr} may obtain superior results if domain knowledge is available to inform the choice of regularization or if a parameter search is performed over both $\lambda$ and $\eta$.

\subsection{Simple Linear Example}

We first consider the methods presented in this work applied to a simple linear regression.  We consider a problem with $\bX \in \bbR^{250 \times 250}$, $\bTheta$ being the identity, and construct random linear maps by setting 25 of 250 coefficients to be Gaussian distributed with unit variance and setting the rest to zero.  Since the othonormal case is explored in the Sections \ref{sec:regularization_methods} and \ref{sec:thresholding_methods} we construct $\bX$ (equivalently $\btheta( \bX)$) to have condition number $\kappa (\bX) = 10^2$ with singular values spread evenly on a log scale between $10^{-2}$ and $0$.  Observations are perturbed by Gaussian noise $\nu \sim \mathcal{N}(0, \sigma^2)$ with $\sigma = 0.1 \, \text{std}(\bTheta \bxi)$.  That is, $\sigma$ is set to ten percent of the standard deviations of the unperturbed output.  The magnitude of the noise is not known by the algorithm and is re-estimated after each iteration.

\begin{figure}
\centering
\includegraphics[width=\textwidth]{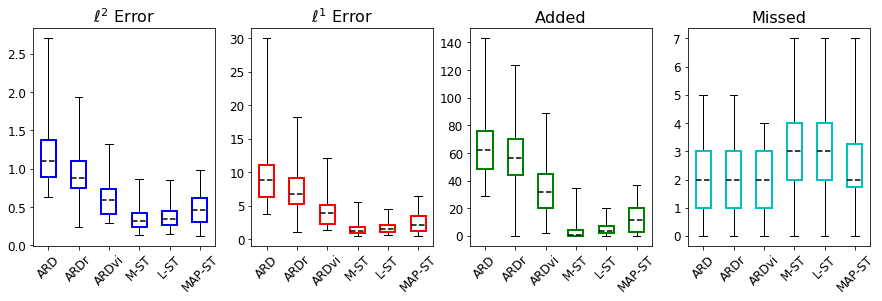}
\caption{Error statistics for 100 trials sparse variants of automatic relevance determination with 250 observations of a 250 dimensional problem having 25 non-zero coefficients.  Boxes indicate inter-quartile range and median error and whiskers show full range of observed values.}
\label{fig:linear_stats}
\end{figure}

We test each of the methods presented in this work for a total of 100 trials, each with random data, true $\bxi$, and noise.  Model selection is performed with $AIC_c$ with a wide range of input parameters.  Four error metrics are tracked; the $\ell^2$ and $\ell^1$ difference between the mean posterior estimate and true $\bxi$ as well as the number of non-zero terms that the learning algorithm adds and misses.  These values are shown in Table \ref{tab:linear_example} and in Fig. \ref{fig:linear_stats}.  Boxes indicate the inter-quartile range and median error across the 100 trials with whiskers indicating maximum and minimum values.  Each method gives far higher $\ell^1$ than $\ell^2$ error indicating these quantities are dominated by the many small terms added by the regression.  However, the thresholding based methods exhibit far lower metric error and number of added terms with only a small increase in the number of missed terms.

\begin{table}
  \begin{center}
    \caption{Mean error for linear system using variations of ARD method.}
    \label{tab:linear_example}
    \begin{tabular}{|l|c|c|c|c|}
      \hline
      Method & $\ell^2$ Error & $\ell^1$ Error & Added & Missed  \\
      \hline
      ARD & $1.2$ & $9.9$ & $65.38$ & $2.15$  \\
      ARDvi & $0.62$ & $4.11$ & $35.13$ & $1.84$  \\
      ARDr & $0.99$ & $7.96$ & $60.44$ & $2.00$  \\
      M-STSBL & $0.35$ & $1.50$ & $3.39$ & $3.21$  \\
      L-STSBL & $0.38$ & $1.76$ & $5.72$ & $2.90$  \\
      MAP-STSBL & $0.47$ & $2.52$ & $12.76$ & $2.74$  \\
      \hline
    \end{tabular}
  \end{center}
\end{table}

\subsection{Interpolation from Few Observations}

We consider fitting a function defined on $T^2 = [0,\pi]^2$ with sparse representation in a Fourier basis.  We let $\bX \in [0, \pi)^{250 \times 2}$ have rows uniformly sampled on $T^2$ and $\btheta : T^2 \to \bbR^{900}$ be the mapping to the basis constructed by the first 30 Fourier modes in each direction so that $\bTheta \in \bbR^{250 \times 900}$.  Similar to the linear example, we set 50 of the 900 coefficients to be Gaussian distributed with unit variance and the rest are set to zero.  Noise is again set to have standard deviation equal to 10 percent of the standard deviation of unperturbed values of $\by$ and the magnitude of the noise is re-estimated after each iteration.

\begin{figure}
\centering
\includegraphics[width=\textwidth]{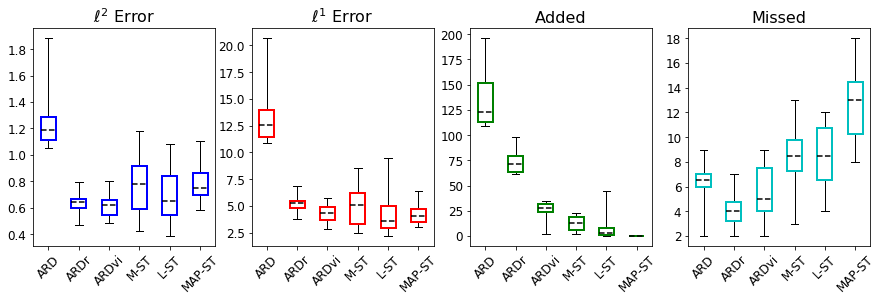}
\caption{Error statistics for 10 trials sparse variants of automatic relevance determination with 250 observations in $T^2$ with two-dimensional Fourier basis $\btheta : T^2 \to \bbR^{900}$ and 50 non-zero coefficients.  Boxes indicate inter-quartile range and median error and whiskers show full range of observed values.}
\label{fig:fourier_stats}
\end{figure}

Results across 10 trials for fitting a function with a sparse Fourier basis are shown in Fig.~\ref{fig:fourier_stats}.  Within each trial we test a wide range of input parameters for each technique and select a model using the $AIC_c$.  Regularization and thresholding techniques all exhibit far lower $\ell^1$ and $\ell^2$ error and include far fewer extraneous terms.  The thresholding methods all show some increase in the number of missed terms.  We expect the increase in false negatives would be lessened if active terms had magnitudes bounded away from zero.

As was the case for the linear example, all regularized methods exhibit lower metric error than unregularized ARD and add substantially fewer extraneous terms.  However, unlike the linear example there is a noteable increase in the number of missing terms using thresholding methods and, contrary to intuition, a decrease in the number of missed terms using the regularization based methods.

\subsection{Equations of Motion for the Lorenz 63 System}

Our first example of applying the techniques to a nonlinear system identification problem is the Lorenz 63 system given by,
\begin{equation}
\begin{aligned}
\dot{x}_1 &= s (x_2-x_1) \\
\dot{x}_2 &= x_1 (\rho - x_3) - x_2 \\
\dot{x}_3 &= x_1x_2 - \beta x_3,
\end{aligned}
\label{eq:L63}
\end{equation}
with the standard set of coefficients $s = 10$, $\rho = 28$ and $\beta = \frac{8}{3}$ \cite{Lorenz1963jas}.  We will follow work by \cite{Brunton2016} for nonlinear system identification and use trajectories from Eq.~\ref{eq:L63} as data $\bX$ and the numerically computed velocity as $\by$.

We construct datasets to test each algorithm by integrating Eq.~\eqref{eq:L63} for 250 steps of length $dt = 0.05$ from an initial condition drawn from $\mathcal{N} \left( (0,0,15), 5^2\bI \right)$ resulting in a times series in $\bbR^{251 \times 3}$.  We add Gaussian noise with standard deviation equal to 1 percent of the standard devition of the time series to get $\bX$ and subsequently compute temporal derivatives $y^{(j)} \approx \dot{x}_j$ using a $6^\text{th}$ order finite difference scheme applied to the noisy time series.  We use the quintic feature map in three variables $\btheta : \bbR^3 \to \bbR^{{5+3\choose 5}}$ given by,
\begin{equation}
\btheta (x_1, x_2, x_3) = \left(1, x_1, x_2, x_3, x_1^2, x_2^2, x_3^2, x_1x_2, \hdots, x_1^4x_3, x_1^5, x_2^5, x_3^5 \right)
\label{eq:quintic_features}
\end{equation}
This gives a matrix $\btheta(\bX ) \in \bbR^{251 \times 56}$.  The system identification problem is then to find sparse solutions to,
\begin{equation}
\dot{x}_j = y^{(j)} = \btheta ( \bx ) \xi^{(j)}
\label{eq:nonlinear_sys_id}
\end{equation}
for each dimension $j = 1,2,3$.  

Note that since noise is added to the data $\bX$ directly rather than to the true $\btheta(\bX)\by$, columns of $\btheta(\bX)$ will be perturbed by nonlinear maps of Gaussian noise.  The error in our polynomial regression will therefore be non-Gaussian, violating the likelihood model we start with in Eq.~\ref{eq:sparse_reg}.  This difference does not significantly affect the regression algorithms but does lead to problems with $AIC_c$ based system identification since the likelihood computed by Eq.~\eqref{eq:neg_log_likelihood} makes assumptions regarding error statistics that do not hold.  We therefore user oracle model selection, choosing the input parameter that yields the minimal number of added and missed terms compared to the true solution.  This of course assumes knowledge of the true solution which would not be the case in an application setting but allows us to focus on comparing sparse regression algorithms rather than on model selection.

\begin{figure}
\centering
\includegraphics[width=\textwidth]{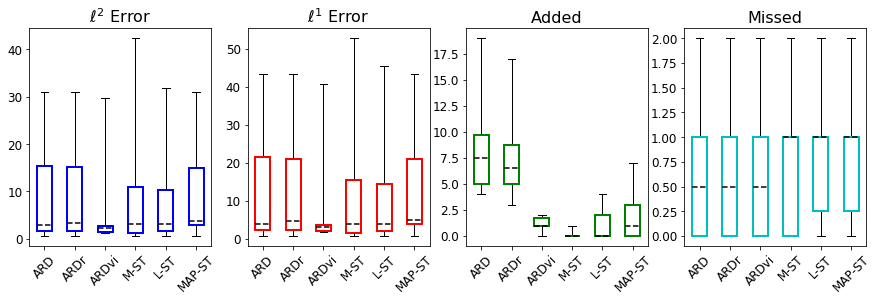}
\caption{Error statistics for 10 trials sparse variants of automatic relevance determination with 251 observed timesteps of a single trajectory of the Lorenz 63 system using a quintic polynomial basis.  Boxes indicate inter-quartile range and median error and whiskers show full range of observed values.}
\label{fig:L63_stats}
\end{figure}

We test each of the methods for ten trials, each using the same length of time series but with different random initial conditions and noise instances.  Figure~\ref{fig:L63_stats} shows error metrics for the coefficients of the learned equations.  Since we are solving three distinct problems, the errors shown in Fig.~\ref{fig:L63_stats} are summed over each of the three dimensions.  Thresholding based methods and variance inflation, learn much sparser models than ARD and ARDr, with ARDvi having no increase in the number of missed terms.  In this case, M-STSBL outperforms both L-STSBL and MAP-STSBL, possibly due to the fact that none of the true coefficients are small.

\subsection{Equations of Motion for the Lorenz 96 System}

We next the consider the higher dimensional Lorenz 96 system given by,
\begin{equation}
\begin{aligned}
\dot{x}_j &= (x_{j+1} - x_{j-2} )x_{j-1} - x_j + F
\end{aligned}
\label{eq:L96}
\end{equation}
with $n=40$ and $F = 16$ \cite{Lorenz1996predictability}.

\begin{figure}
\centering
\includegraphics[width=\textwidth]{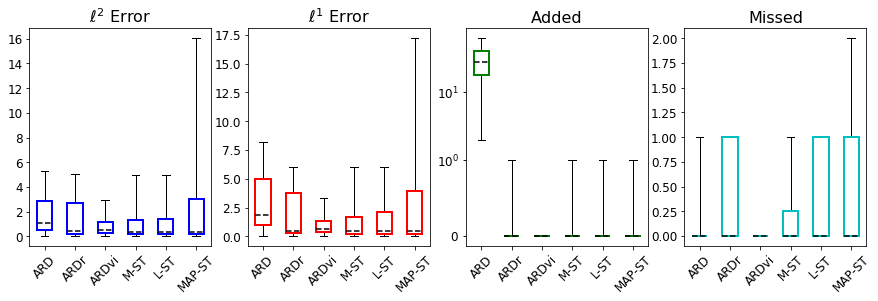}
\caption{Error statistics across 40 dimensions using sparse variants of automatic relevance determination with 201 observed timesteps of a single trajectory of the Lorenz 96 system using a quadratic polynomial basis.  Boxes indicate inter-quartile range and median error and whiskers show full range of observed values.}
\label{fig:L96_stats}
\end{figure}

We construct a dataset to test each algorithm by integrating Eq.~\eqref{eq:L96} with $dt = 0.05$ from an initial condition $x_j = \exp (-\frac{1}{16} (j-20)^2)$ resulting in a times series in $\bbR^{200 \times 40}$.  We add Gaussian noise with standard deviation equal to 1 percent of the standard devition of the time series to get $\bX$ and subsequently compute temporal derivatives $y^{(j)} \approx \dot{x}_j$ using a $6^\text{th}$ order finite difference scheme applied to the noisy time series.  We use the quadratic feature map in 40 variables $\btheta : \bbR^{40} \to \bbR^{{2+40\choose 2}}$ given by,
\begin{equation}
\btheta (x_1, x_2, \hdots , x_{40}) = \left(1, x_1, x_2, \hdots, x_{40} , x_1^2 , x_2^2, \hdots, x_{40}^2, x_1x_2, x_1x_3, \hdots , x_{39}x_{40} \right)
\label{eq:quadratic_features}
\end{equation}
This gives a matrix $\btheta(\bX ) \in \bbR^{201 \times 861}$.  We solve for the equations of motion just as we did in the Lorenz 63 case.  Model selection is again performed assuming full knowledge of the true sparsity pattern.

We test each of the methods on a single trial across each of the 40 dimensions.  Figure~\ref{fig:L96_stats} shows error metrics for the coefficients of the learned equations across the 40 dimensions.
Each of the proposed techniques learns a more sparse set of coefficients than ARD with the modal number of added terms for each of the proposed methods being zero.  However, metric error is not improved significantly and in the case of MAP-STSBL contains outlier values with substantially increased error where the algorithm failed to include the forcing term $F=16$.  This indicates that extraneous terms in the ARD estimate were generally small.  The modal number of missed terms for each method is zero, but all methods except ARDvi added a single term in a some fraction of the dimensions and MAP-STSBL occasionally missing two.

\subsection{Equations of Motion for the Kuramoto-Sivashinsky Equation}


We also test each of the sparse regression methods considered in this work on the Kuramoto Sivashinsky (KS) equation.  The KS equation, given by, 
\begin{equation}
u_t + uu_x + u_{xx} + u_{xxxx} = 0,
\label{eq:KS}
\end{equation}
is often used as model for deterministic spatiotemporal chaos and has proved a challenging case for other sparse regression methods \cite{raissi2018deep, rudy2019data}.

We use the ETDRK4 method developed in \cite{kassam2005fourth} to solve the Kuramoto Sivashinsky equation on the domain $(x,t) \in [0,32 \pi ] \times [0,150]$ with periodic boundary conditions, initial condition $x$, timesteps $dt = 0.14$ and spatial discretization $dx = 32\pi/512$.  We add artificial noise to the numerical solution with standard deviation equal to 0.1 percent of the standard deviation of the data.  This small magnitude is consistent with previous published works in system identification for the KS equation.  Numerical differentiation with respect to time and for the first four spatial derivatives is done by applying sixth order finite difference schemes directly to the noisy data.  We take $y$ to be $u_t$ reshaped into a vector and $\btheta$ to be the set of powers of $u$ up to $4$ multiplied by spatial derivatives up to fourth order so that
\begin{equation}
\btheta(\bX) = \btheta ( u(x,t) ) = \left(1,  u, \hdots u^2, u_x, uu_x, \hdots u^4 u_{xxxx}  \right).
\label{eq:KS_features}
\end{equation}

\begin{figure}
\centering
\includegraphics[width=\textwidth]{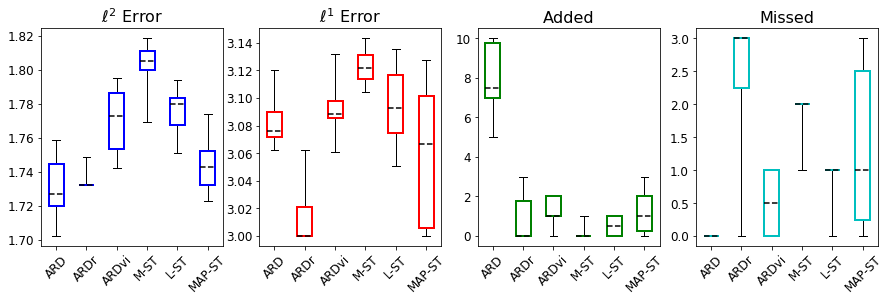}
\caption{Error statistics for 10 trials of usiong 2500 randomly selected rows for finding the Kuramoto Sivashinsky equation with standard length scales.  Boxes indicate inter-quartile range and median error and whiskers show full range of observed values.}
\label{fig:KS_stats}
\end{figure}

With the given discretization of Eq. \eqref{eq:KS}, the feature map \eqref{eq:KS_features} gives $\by \in \bbR^{(1024 \cdot 512) \times 1}$ and $\btheta(\bX) \in \bbR^{(1024 \cdot 512) \times 25}$.  Each iteration of Alg.~\ref{alg:ARD} requires storing and inverting $\bSigma_y \in \bbR^{m \times m}$ where $m 1024 \cdot 512 $ is the number of observations.  Allocating memory for and working with $\bSigma_y$ in this case would be problematic on many standard computers.  We instead observe a small fraction of the data through random projections, exploiting the simple fact that,
\begin{equation}
\by = \btheta(\bX) \bxi \rightarrow \bC \by = \bC \btheta(\bX) \bxi,
\label{eq:subsampling}
\end{equation}
for any matrix $\bC$.  We take each column of $\bC$ to be a unit direction vector sampled uniformly and without replacement from $\bbR^{1024 \cdot 512}$ so that we are simply sampling rows from the full linear system.  To test the effectiveness of each algorithm we take 10 different samples of size 2500 and solve the linear system for each one.  Figure \ref{fig:KS_stats} summarized the error of each of the proposed regressions applied to the 10 random subsets given by Eq. \ref{eq:subsampling}.

While none of the proposed methods perform well for the task of identifying the Kuramoto Sivashinsky equation from data, the proposed sparse methods do learn more parsimonious models.  This comes at the cost of higher $\ell^2$ error and a significantly increased number of missing terms.  The modal number of missed terms for ARDr is in fact all three of the non-zero terms.  While these results are dissapointing, they are also unsurprising.  The Kuramoto Sivashinsky equation has proved challenging for past system identification methods \cite{rudy2019data}.  This example showcases some of the limitations of the methodology proposed in this work and the continuing difficulty of sparse regression based methods, both classical and Bayesian, for system identification.

\section{Discussion}\label{sec:discussion}

We have presented several techniques for learning sparse Bayesian methods that build on Automatic Relevance Determination to achieve greater levels of parsimony in the resulting linear model.
These methods may be classified in two families; regularization based methods including variance inflation and regularization of $\bgamma$, which find the variance coefficients $\bgamma$ as the fixed point of a single application of an iterative algorithm, and thresholding based methods, which alternate between solving a smooth optimization problem and simplifying the model via thresholding extraneous terms.  
For the latter class we tested magnitude based thresholding based on the mean posterior estimate of $\bxi$, as well as a likelihood based threshold using the posterior distribution, and adjusting the prior on $\bxi$ to find an alternative probabilistic threshold.

For each of these algorithms, we have derived probabilistic estimates for the number of false positive and false negative active terms in the orthogonal case.  
While most practical problems involve non-orthonogonal matrices, these estimates can be taken as guides for the behavior of the algorithms as regularization or thresholding parameters change.

A significant barrier to use of the proposed class of sparse regression methods on many problems is the computational complexity.  Each iteration of Alg.~\ref{alg:ARDr} requires computing the inverse of an $m \times m$ matrix, where $m$ is the number of samples available.  Future work could explore low-rank approximations of this step, but in the current work this was a computational bottleneck and forced us to only consider small problems.  Subsampling approaches such as those in \cite{zhang2019robust} might also be useful for large datasets.

We stress that this work does not attempt to demonstrate the superiority of any of the proposed methods for the subset selection problem in sparse Bayesian regression.  Ultimately, if one desires a level of sparsity beyond that provided by standard ARD, a choice of additional assumptions should be made with respect to the context of the problem being considered.  We have outlined the assumptions that lead to each of the proposed algorithms and demonstrated their accuracy both analytically on orthogonal linear systems as a canonical test case and empirically on several more complicated problems.  In application settings, model selection could be performed both over parameter values for each algorithm as well as between algorithms to determine a final result.

\section*{Acknowledgments}
This material is based upon work supported by the National Science Foundation under Award
No. 1902972, the Army Research Office (Grant No. W911NF-17-1-0306), and a MathWorks Faculty Research Innovation Fellowship.

\bibliographystyle{plain}
\bibliography{references}

\begin{thebibliography}{10}

\bibitem{akaike1974new}
Hirotugu Akaike.
\newblock A new look at the statistical model identification.
\newblock {\em IEEE transactions on automatic control}, 19(6):716--723, 1974.

\bibitem{almomani2020entropic}
Abd AlRahman~R AlMomani, Jie Sun, and Erik Bollt.
\newblock How entropic regression beats the outliers problem in nonlinear
  system identification.
\newblock {\em Chaos: An Interdisciplinary Journal of Nonlinear Science},
  30(1):013107, 2020.

\bibitem{babacan2009bayesian}
S~Derin Babacan, Rafael Molina, and Aggelos~K Katsaggelos.
\newblock Bayesian compressive sensing using laplace priors.
\newblock {\em IEEE Transactions on image processing}, 19(1):53--63, 2009.

\bibitem{bishop2006pattern}
Christopher~M Bishop.
\newblock {\em Pattern recognition and machine learning}.
\newblock springer, 2006.

\bibitem{blumensath2009iterative}
Thomas Blumensath and Mike~E Davies.
\newblock Iterative hard thresholding for compressed sensing.
\newblock {\em Applied and computational harmonic analysis}, 27(3):265--274,
  2009.

\bibitem{Bongard2007pnas}
Josh Bongard and Hod Lipson.
\newblock Automated reverse engineering of nonlinear dynamical systems.
\newblock {\em PNAS}, 104(24):9943--9948, 2007.

\bibitem{boyd2011distributed}
Stephen Boyd, Neal Parikh, Eric Chu, Borja Peleato, Jonathan Eckstein, et~al.
\newblock Distributed optimization and statistical learning via the alternating
  direction method of multipliers.
\newblock {\em Foundations and Trends{\textregistered} in Machine learning},
  3(1):1--122, 2011.

\bibitem{Brunton2016}
S.~L. Brunton, J.~L. Proctor, and J.~N. Kutz.
\newblock Discovering governing equations from data by sparse identification of
  nonlinear dynamical systems.
\newblock {\em PNAS}, 113(15):3932--3927, 2016.

\bibitem{cavanaugh1997unifying}
Joseph~E Cavanaugh.
\newblock Unifying the derivations for the akaike and corrected akaike
  information criteria.
\newblock {\em Statistics \& Probability Letters}, 33(2):201--208, 1997.

\bibitem{efron2004least}
Bradley Efron, Trevor Hastie, Iain Johnstone, Robert Tibshirani, et~al.
\newblock Least angle regression.
\newblock {\em The Annals of statistics}, 32(2):407--499, 2004.

\bibitem{fan2001variable}
Jianqing Fan and Runze Li.
\newblock Variable selection via nonconcave penalized likelihood and its oracle
  properties.
\newblock {\em Journal of the American statistical Association},
  96(456):1348--1360, 2001.

\bibitem{fuentes2019efficient}
R~Fuentes, N~Dervilis, K~Worden, and EJ~Cross.
\newblock Efficient parameter identification and model selection in nonlinear
  dynamical systems via sparse bayesian learning.
\newblock In {\em Journal of Physics: Conference Series}, volume 1264, page
  012050. IOP Publishing, 2019.

\bibitem{graves2011practical}
Alex Graves.
\newblock Practical variational inference for neural networks.
\newblock In {\em Advances in neural information processing systems}, pages
  2348--2356, 2011.

\bibitem{kassam2005fourth}
Aly-Khan Kassam and Lloyd~N Trefethen.
\newblock Fourth-order time-stepping for stiff pdes.
\newblock {\em SIAM Journal on Scientific Computing}, 26(4):1214--1233, 2005.

\bibitem{kim2017causation}
Pileun Kim, Jonathan Rogers, Jie Sun, and Erik Bollt.
\newblock Causation entropy identifies sparsity structure for parameter
  estimation of dynamic systems.
\newblock {\em Journal of Computational and Nonlinear Dynamics}, 12(1), 2017.

\bibitem{li2002bayesian}
Yi~Li, Colin Campbell, and Michael Tipping.
\newblock Bayesian automatic relevance determination algorithms for classifying
  gene expression data.
\newblock {\em Bioinformatics}, 18(10):1332--1339, 2002.

\bibitem{Lorenz1963jas}
Edward~N Lorenz.
\newblock Deterministic nonperiodic flow.
\newblock {\em J. Atmos. Sciences}, 20(2):130--141, 1963.

\bibitem{Lorenz1996predictability}
Edward~N Lorenz.
\newblock Predictability: A problem partly solved.
\newblock In {\em Proc. Seminar on predictability}, volume~1, 1996.

\bibitem{murphy_ml}
Kevin~P. Murphy.
\newblock {\em Machine Learning: A Probabilistic Perspective}.
\newblock The MIT Press, 2012.

\bibitem{neal2012bayesian}
Radford~M Neal.
\newblock {\em Bayesian learning for neural networks}, volume 118.
\newblock Springer Science \& Business Media, 2012.

\bibitem{niven2020bayesian}
Robert~K Niven, Ali Mohammad-Djafari, Laurent Cordier, Markus Abel, and Markus
  Quade.
\newblock Bayesian identification of dynamical systems.
\newblock {\em Multidisciplinary Digital Publishing Institute Proceedings},
  33(1):33, 2020.

\bibitem{oh2008bayesian}
Chang~Kook Oh, James~L Beck, and Masumi Yamada.
\newblock Bayesian learning using automatic relevance determination prior with
  an application to earthquake early warning.
\newblock {\em Journal of Engineering Mechanics}, 134(12):1013--1020, 2008.

\bibitem{parikh2014proximal}
Neal Parikh, Stephen Boyd, et~al.
\newblock Proximal algorithms.
\newblock {\em Foundations and Trends{\textregistered} in Optimization},
  1(3):127--239, 2014.

\bibitem{quade2016prediction}
Markus Quade, Markus Abel, Kamran Shafi, Robert~K Niven, and Bernd~R Noack.
\newblock Prediction of dynamical systems by symbolic regression.
\newblock {\em Physical Review E}, 94(1):012214, 2016.

\bibitem{raissi2018deep}
Maziar Raissi.
\newblock Deep hidden physics models: Deep learning of nonlinear partial
  differential equations.
\newblock {\em The Journal of Machine Learning Research}, 19(1):932--955, 2018.

\bibitem{rasmussen2003gaussian}
Carl~Edward Rasmussen.
\newblock Gaussian processes in machine learning.
\newblock In {\em Summer School on Machine Learning}, pages 63--71. Springer,
  2003.

\bibitem{rudy2019data}
Samuel Rudy, Alessandro Alla, Steven~L Brunton, and J~Nathan Kutz.
\newblock Data-driven identification of parametric partial differential
  equations.
\newblock {\em SIAM Journal on Applied Dynamical Systems}, 18(2):643--660,
  2019.

\bibitem{schaeffer2017learning}
Hayden Schaeffer.
\newblock Learning partial differential equations via data discovery and sparse
  optimization.
\newblock In {\em Proc. R. Soc. A}, volume 473, page 20160446. The Royal
  Society, 2017.

\bibitem{Schmidt2009science}
Michael Schmidt and Hod Lipson.
\newblock Distilling free-form natural laws from experimental data.
\newblock {\em Science}, 324(5923):81--85, 2009.

\bibitem{tan2012automatic}
Vincent~YF Tan and C{\'e}dric F{\'e}votte.
\newblock Automatic relevance determination in nonnegative matrix factorization
  with the/spl beta/-divergence.
\newblock {\em IEEE Transactions on Pattern Analysis and Machine Intelligence},
  35(7):1592--1605, 2012.

\bibitem{tibshirani1996regression}
Robert Tibshirani.
\newblock Regression shrinkage and selection via the lasso.
\newblock {\em Journal of the Royal Statistical Society: Series B
  (Methodological)}, 58(1):267--288, 1996.

\bibitem{tipping2001sparse}
Michael~E Tipping.
\newblock Sparse bayesian learning and the relevance vector machine.
\newblock {\em Journal of machine learning research}, 1(Jun):211--244, 2001.

\bibitem{tipping2003fast}
Michael~E Tipping, Anita~C Faul, et~al.
\newblock Fast marginal likelihood maximisation for sparse bayesian models.
\newblock In {\em AISTATS}, 2003.

\bibitem{tran2017exact}
Giang Tran and Rachel Ward.
\newblock Exact recovery of chaotic systems from highly corrupted data.
\newblock {\em Multiscale Modeling \& Simulation}, 15(3):1108--1129, 2017.

\bibitem{wipf2004basis}
D.~P. {Wipf} and B.~D. {Rao}.
\newblock Sparse bayesian learning for basis selection.
\newblock {\em IEEE Transactions on Signal Processing}, 52(8):2153--2164, 2004.

\bibitem{wipf2008new}
David~P Wipf and Srikantan~S Nagarajan.
\newblock A new view of automatic relevance determination.
\newblock In {\em Advances in neural information processing systems}, pages
  1625--1632, 2008.

\bibitem{wu2008coordinate}
Tong~Tong Wu, Kenneth Lange, et~al.
\newblock Coordinate descent algorithms for lasso penalized regression.
\newblock {\em The Annals of Applied Statistics}, 2(1):224--244, 2008.

\bibitem{wu2012dual}
Yi~Wu and David~P Wipf.
\newblock Dual-space analysis of the sparse linear model.
\newblock In {\em Advances in Neural Information Processing Systems}, pages
  1745--1753, 2012.

\bibitem{yang2020bayesian}
Yibo Yang, Mohamed~Aziz Bhouri, and Paris Perdikaris.
\newblock Bayesian differential programming for robust systems identification
  under uncertainty.
\newblock {\em arXiv preprint arXiv:2004.06843}, 2020.

\bibitem{yuan2019machine}
Ye~Yuan, Junlin Li, Liang Li, Frank Jiang, Xiuchuan Tang, Fumin Zhang, Sheng
  Liu, Jorge Goncalves, Henning~U Voss, Xiuting Li, et~al.
\newblock Machine discovery of partial differential equations from
  spatiotemporal data.
\newblock {\em arXiv preprint arXiv:1909.06730}, 2019.

\bibitem{zhang2019convergence}
Linan Zhang and Hayden Schaeffer.
\newblock On the convergence of the sindy algorithm.
\newblock {\em Multiscale Modeling \& Simulation}, 17(3):948--972, 2019.

\bibitem{zhang2018robust}
Sheng Zhang and Guang Lin.
\newblock Robust data-driven discovery of governing physical laws with error
  bars.
\newblock {\em Proceedings of the Royal Society A: Mathematical, Physical and
  Engineering Sciences}, 474(2217):20180305, 2018.

\bibitem{zhang2019robust}
Sheng Zhang and Guang Lin.
\newblock Robust data-driven discovery of governing physical laws using a new
  subsampling-based sparse bayesian method to tackle four challenges (large
  noise, outliers, data integration, and extrapolation).
\newblock {\em arXiv preprint arXiv:1907.07788}, 2019.

\bibitem{zhang2009greedy}
Tong Zhang.
\newblock Adaptive forward-backward greedy algorithm for sparse learning with
  linear models.
\newblock In D.~Koller, D.~Schuurmans, Y.~Bengio, and L.~Bottou, editors, {\em
  Advances in Neural Information Processing Systems 21}, pages 1921--1928.
  Curran Associates, Inc., 2009.

\bibitem{zheng2018unified}
Peng Zheng, Travis Askham, Steven~L Brunton, J~Nathan Kutz, and Aleksandr~Y
  Aravkin.
\newblock A unified framework for sparse relaxed regularized regression: Sr3.
\newblock {\em IEEE Access}, 7:1404--1423, 2018.

\end{thebibliography}

\newpage
\section*{Appendix A: Proof of Equation \eqref{eq:convex_loss_equiv}}

We show that $\by^T\bSigma_y^{-1}\by = \underset{\bxi}{min} \frac{1}{\sigma^2} \|\by - \bTheta \bxi \|_2^2 + \bxi^T\bGamma^{-1}\bxi$.  Applying the Woodbury identity to $\bSigma_y^{-1}$ gives,
\begin{align*}
\by^T\bSigma_y^{-1}\by &= \by^T (\sigma^{2} \bI + \bTheta \bGamma \bTheta^T)^{-1} \by \\
&= \by^T (\sigma^{-2} \bI - \sigma^{-4} \bTheta (\bGamma^{-1} + \sigma^{-2} \bTheta^T\bTheta)^{-1} \bTheta^T) \by \\
&= \by^T (\sigma^{-2} \bI - \sigma^{-4} \bTheta \bSigma_\xi \bTheta^T) \by,
\end{align*}
On the other hand,
\begin{align*}
\underset{\bxi}{min} \frac{1}{\sigma^2} &\|\by - \bTheta \bxi \|_2^2 + \bxi^T\bGamma^{-1}\bxi = \frac{1}{\sigma^2} \|\by - \bTheta \bmu_\xi \|_2^2 + \bmu_\xi^T\bGamma^{-1}\bmu_\xi \\
&= \frac{1}{\sigma^2} \|\by - \sigma^{-2} \bTheta \bSigma_\xi \bTheta^T \by \|_2^2 + \sigma^{-4} \by^T \bTheta \bSigma_\xi \bGamma^{-1} \bSigma_\xi \bTheta^T \by \\
&= \frac{1}{\sigma^2} \by^T\by - \frac{2}{\sigma^4}\by^T \bTheta \bSigma_\xi \bTheta^T \by + \frac{1}{\sigma^6} \by^T\bTheta \bSigma_\xi \bTheta^T \bTheta \bSigma_\xi \bTheta^T \by + \frac{1}{\sigma^4} \by^T \bTheta \bSigma_\xi \bGamma^{-1} \bSigma_\xi \bTheta^T \by \\
&= \frac{1}{\sigma^2} \by^T\by + \frac{1}{\sigma^4} \by^T \bTheta \bSigma_\xi \left( -2\bI + \left( \frac{1}{\sigma^2}\bTheta^T\bTheta + \bGamma^{-1} \right) \bSigma_\xi   \right) \bTheta^T \by\\
&= \frac{1}{\sigma^2} \by^T\by + \frac{1}{\sigma^4} \by^T \bTheta \bSigma_\xi \left( -2\bI + \bSigma_\xi^{-1} \bSigma_\xi   \right) \bTheta^T \by\\
&= \frac{1}{\sigma^2} \by^T\by - \frac{1}{\sigma^4} \by^T \bTheta \bSigma_\xi \bTheta^T \by\\
&= \by^T (\sigma^{-2} \bI - \sigma^{-4} \bTheta \bSigma_\xi \bTheta^T) \by,
\end{align*}

\section*{Appendix B: Converse of Equation \eqref{eq:ARDr_xi0_cond}}

In this section we show that,
\begin{equation}
\left| \xi_i - \rho_i^{-1} \bTheta_i^T\bnu \right| \leq \sqrt{ \rho_i^{-1} \sigma^{2} + \lambda\rho_i^{-2} \sigma^{4}} \Rightarrow \xi_i^* = 0
\end{equation}
From the above inequality and the KKT stationarity condition for the $\xi$ we have,
\begin{equation}
\begin{aligned}
\sqrt{ \rho_i^{-1} \sigma^{2} + \lambda\rho_i^{-2} \sigma^{4}} &\geq \left| \xi_i^* + \rho_i^{-1}\sigma^2 \sqrt{c_i^*} sgn (\xi_i^*)  \right| \\
&= \left| \xi_i^* \right| + \rho_i^{-1}\sigma^2 \sqrt{c_i^*}  \\
\therefore \sqrt{c_i^*} &\leq \sqrt{\lambda + \rho_i\sigma^{-2}} -\rho_i\sigma^{-2}|\xi_i^*| \\
\text{ and } |\xi_i^*| &\leq \sqrt{ \rho_i^{-1} \sigma^{2} + \lambda\rho_i^{-2} \sigma^{4}} - \rho_i^{-1}\sigma^2 \sqrt{c_i^*}
\label{eq:KKT_inequality}
\end{aligned}
\end{equation}
From Eq.~\eqref{eq:ARDr_root_c*} $\sqrt{c_i^*}$ is given by the positive valued zero of following cubic,
\begin{equation}
\psi(\omega) = \rho_i^{-1}\sigma^2 \omega^3 + |\xi_i^*| \omega^2 - \left( \lambda\rho_i^{-1} \sigma^2 +1 \right)  \omega -\lambda |\xi_i^*|
\label{eq:psi_cubic}
\end{equation}
where $\omega = \sqrt{c_i^*}$ to simplify notation.  Note that $\psi(0) \leq 0$ with equality only if $\lambda$ or $\xi_i^* = 0$, $\psi '(0) < 0$, and the coefficient on the cubic term is positive.  This suffices to show there is a unique positive zero of $\psi$.  We also know that $\sqrt{c_i^*}$ is greater than the larger of the two zeros of $\psi'(\omega)$ given by,
\begin{equation}
\sqrt{c_i^*} > \omega^+ = \frac{\rho_i}{3\sigma^2} \left(  -|\xi_i^*| + \sqrt{ {\xi_i^*}^2 + 3\sigma^2\rho_i^{-1}\left(1+\lambda\sigma^2\rho_i^{-1}\right)}  \right). \label{eq:omega_plus}
\end{equation}
Substituting the lower bound for $\sqrt{c_i^*}$ given by Eq.~\eqref{eq:omega_plus} into Eq.~\eqref{eq:KKT_inequality} gives,
\begin{equation}
\begin{aligned}
|\xi_i^*| &< \sqrt{ \rho_i^{-1} \sigma^{2} + \lambda\rho_i^{-2} \sigma^{4}} -\frac{1}{3} \left(  -|\xi_i^*| + \sqrt{ {\xi_i^*}^2 + 3\sigma^2\rho_i^{-1}\left(1+\lambda\sigma^2\rho_i^{-1}\right)}  \right) \\
\frac{2 |\xi_i^*|}{3} &< \sqrt{ \rho_i^{-1} \sigma^{2} + \lambda\rho_i^{-2} \sigma^{4}} -\frac{1}{3} \sqrt{ {\xi_i^*}^2 + 3\sigma^2\rho_i^{-1}\left(1+\lambda\sigma^2\rho_i^{-1}\right)} \\
|\xi_i^*| &< \frac{3}{2} \sqrt{ \rho_i^{-1} \sigma^{2} + \lambda\rho_i^{-2} \sigma^{4}} - \frac{1}{2} \sqrt{ {\xi_i^*}^2 + 3\sigma^2\rho_i^{-1}\left(1+\lambda\sigma^2\rho_i^{-1}\right)} \\
&\leq \frac{3}{2} \sqrt{ \rho_i^{-1} \sigma^{2} + \lambda\rho_i^{-2} \sigma^{4}} - \frac{1}{2} \sqrt{  3\sigma^2\rho_i^{-1}\left(1+\lambda\sigma^2\rho_i^{-1}\right)}  \\
&= \frac{3-\sqrt{3}}{2} \sqrt{ \rho_i^{-1} \sigma^{2} + \lambda\rho_i^{-2} \sigma^{4}}
\end{aligned}
\end{equation}
From Eq.~\eqref{eq:KKT_inequality} we know $\sqrt{c_i^*} \leq \sqrt{\lambda + \rho_i\sigma^{-2}} -\rho_i\sigma^{-2}|\xi_i^*|$.  Since $\sqrt{c_i^*}$ is the greatest zero of $\psi$ and the cubic coefficient is positive,
\begin{equation}
\begin{aligned}
0 &\leq \psi \left( \sqrt{\lambda + \rho_i\sigma^{-2}} -\rho_i\sigma^{-2}|\xi_i^*| \right)  \\
&= \frac{\rho_i |\xi_i^*| }{\sigma^2} \left( |\xi_i^*| \sqrt{\lambda + \rho_i\sigma^{-2}} - 2\lambda \rho_i^{-1}\sigma^2 -1  \right) \\
&\leq \frac{\rho_i |\xi_i^*| }{\sigma^2} \left( \frac{3-\sqrt{3}}{2} (1 + \lambda\rho_i^{-1}\sigma^2) - 2\lambda\rho_i^{-1} \sigma^2 -1  \right) \\
&= \frac{\rho_i^{-1} |\xi_i^*| }{\sigma^2} \left( \frac{-1-\sqrt{3}}{2} \lambda\rho_i^{-1}\sigma^2 + \frac{1-\sqrt{3}}{2}\right)
\end{aligned}
\end{equation}
Note that the quantity inside the parentheses is strictly less than zero.  Therefore, for the inequality to hold, $|\xi_i^*| = 0$.

\section*{Appendix C: Comparrison Between L-STSBL and MAP-STSBL}

The thresholding operations introduced for algorithms \ref{alg:L-STSBL} and \ref{alg:MAP-STSBL} bear some similarities but differ in an important manner with regards to how they treat the posterior marginal variance of $\xi_i$.  The thresholding criteria for $\xi_i \to 0$ in Alg. \ref{alg:L-STSBL} given threshold $\tau_0$ is,
\begin{equation}
h_L(\mu_{\xi,i}, \Sigma_{\xi,ii}) =  \dfrac{1}{\sqrt{2\pi \Sigma_{\xi, ii}}}\exp\left( \frac{-\mu_{\xi,i}^2}{2 \Sigma_{\xi,ii}} \right) > \tau_0,
\end{equation}
while for Alg \ref{alg:MAP-STSBL} it is,
\begin{equation}
h_{MAP}(\mu_{\xi,i}, \Sigma_{\xi,ii})  = \frac{\mu_{\xi,i}^2}{2 \Sigma_{\xi,ii}} < \tau_1,
\end{equation}
or equivalently,
\begin{equation}
exp(h_{MAP}(-\mu_{\xi,i}, \Sigma_{\xi,ii})) = \exp\left( \frac{-\mu_{\xi,i}^2}{2 \Sigma_{\xi,ii}} \right) > e^{-\tau_1} = \tau_2.
\end{equation}
The two criteria are related by,
\begin{equation}
h_L(\mu_{\xi,i}, \Sigma_{\xi,ii}) = \dfrac{exp(-h_{MAP}(\mu_{\xi,i}, \Sigma_{\xi,ii}))}{\sqrt{2\pi \Sigma_{\xi, ii}}}.
\end{equation}
This highlights the difference in assumptions between the two methods.  In both algorithms, high uncertainty relative to coefficient magnitude indicates a greater chance of pruning.  However, this effect is slightly lessened in Algorithm \ref{alg:L-STSBL}.  Coefficients with low uncertainty relative to their magnitude are unlikely to be pruned using either method but the likelihood is higher using \ref{alg:L-STSBL}.
\end{document}